# GOOSE Algorithm: A Powerful Optimization Tool for Real-World Engineering Challenges and Beyond

Rebwar Khalid Hamad[1][0000-0002-3490-3721] and Tarik A. Rashid[2][0000-0002-8661-258X]*

[1]Department of Information Systems Engineering, Erbil Technical Engineering College, Erbil Polytechnic University, Erbil, Iraq

[2]Computer Science and Engineering Department, University of Kurdistan Hewler, Erbil, Iraq

Correspondence email: tarik.ahmed@ukh.edu.krd

**Abstract**

This study proposes the GOOSE algorithm as a novel metaheuristic algorithm based on the goose's behaviour during rest and foraging. The goose stands on one leg and keeps his balance to guard and protect other individuals in the flock. The GOOSE algorithm is benchmarked on 19 well-known benchmark test functions, and the results are verified by a comparative study with genetic algorithm (GA), particle swarm optimization (PSO), dragonfly algorithm (DA), and fitness dependent optimizer (FDO). In addition, the proposed algorithm is tested on 10 modern benchmark functions, and the gained results are compared with three recent algorithms, such as the dragonfly algorithm, whale optimization algorithm (WOA), and salp swarm algorithm (SSA). Moreover, the GOOSE algorithm is tested on 5 classical benchmark functions, and the obtained results are evaluated with six algorithms, such as fitness dependent optimizer (FDO), FOX optimizer, butterfly optimization algorithm (BOA), whale optimization algorithm, dragonfly algorithm, and chimp optimization algorithm (ChOA). The achieved findings attest to the proposed algorithm's superior performance compared to the other algorithms that were utilized in the current study. The technique is then used to optimize Welded beam design and Economic Load Dispatch Problems, pressure vessel design problems, and the Pathological IgG Fraction in the Nervous System, four renowned real-world challenges. The outcomes of the engineering case studies illustrate how well the suggested approach can optimize issues that arise in the real world.

1. ## Introduction

Since computers came along, the main goal has been to find the best solution. At the government's wartime communications center, Alan Turing spent the majority of his time between 1939 and 1945 perfecting the German enciphering machine Enigma and conducting other cryptological research. Turing became the leading scientist with specific responsibility for deciphering the U-boat transmissions after making a distinctive logical breakthrough in the decoding of the Enigma. As a result, he rose to prominence in Anglo-American relations and was exposed to the most cutting-edge electrical technologies of the time. From time to date thousands of algorithms have been designed for different kinds of goals, notably optimization problems. The optimization Problems are solved using a metaheuristic technique. Numerous facets of everyday living might suffer from optimization issues. In general, there are two types of optimisation algorithms: classical and evolutionary. Quadratic programming and gradient-based algorithms are examples of conventional algorithms. Heuristic or metaheuristic algorithms and several hybrid methods are examples of evolutionary algorithms. In recent years, the employment of metaheuristic algorithms has become common practice to resolve modern-day real-world optimization problems, which cannot be resolved by conventional mathematical methods.

Meta-heuristics may or may not take cues from nature. Evolutionary algorithms, physics-based algorithms, swarm-based algorithms, and human-based algorithms are the four main types of nature-inspired meta-heuristic algorithms. In recent years, modern metaheuristic algorithms have begun to prove their efficacy in solving challenging optimisation issues and even NP-hard problems(Rahman et al., 2021).

Given the evolution of many metaheuristic algorithms over the last several decades, classifying them into one of four broad groups is possible. The first group, "Evolutionary Algorithms" [EAs] includes algorithms like the Genetic Algorithm (Holland, 1975), Differential Evolution (DE)(Storn & Price, 1997), Tabu Search(TS) (Glover, 1989), and Biogeography-Based Optimizer (BBO) (Simon, 2008), An improved stochastic fractal search algorithm (ISFS) (Çelik, 2020b). The second group consists of the algorithms created based on "Swarm Intelligence" [SIs] such as Particle Swarm Optimization(PSO) (Kennedy & Eberhart, 1995), Ant Colony Optimization(ACO) (Dorigo & Di Caro, 1999), Firefly Algorithm(FA) (X.-S. Yang, 2008), (GaTSa) (Zanchettin et al., 2011). The "Physics-Inspired Algorithms" [PIAs] are the third group including Harmony Search(HS) (Geem et al., 2001), Big-Bang Big-Crunch (BBBC) (Erol & Eksin, 2006), Gravitational Search Algorithm (GSA) (Rashedi et al., 2009). The algorithms in the last category have been developed based on information on both human and animal lifestyles such as Simulated Annealing(SA) (Kirkpatrick et al., 1983), Evolutionary Algorithm(EA)(Bäck & Schwefel, 1993), Cultural Algorithm

(CA) (Sebald & Fogel, 1994), Artificial Bee Colony (ABC) (Karaboga & Basturk, 2007), Monkey Algorithm (MA) (Zhao & Tang, 2008), Bat Algorithm (BA) (X. S. Yang, 2010), Teaching-Learning-Based Optimization (TLBO)(Rao et al., 2011), Optimization Technique courses (OT) (Precup et al., 2021), Bacterial Colony Optimization(BCO) (Niu & Wang, 2012), Krill Herd Algorithm(KHA) (Gandomi & Alavi, 2012), Cuckoo Search(CS) (Gandomi et al., 2013), Ant Nesting Algorithm (ANA) (Rashid et al., 2021), Grey Wolf Optimizer (GWO) (Mirjalili et al., 2014), Dragonfly algorithm (Mirjalili, 2016), Genetic Algorithm-based Simulated Annealing (GASA) (Kaplan & Çelik, 2018), Donkey and smuggler optimization algorithm (DSO) (Shamsaldin et al., 2019), Fitness Dependent Optimizer (Abdullah & Ahmed, 2019), Child Drawing Development Optimization Algorithm Based on Child's Cognitive Development (CDDO) (Abdulhameed & Rashid, 2022), Improved Sooty Tern Optimization Algorithm (mSTOA) (Houssein et al., 2023), It presents an innovative reference tracking control strategy that combines the Grey Wolf Optimizer (GWO) algorithm with the Actor-Critic Reinforcement Learning (RL) framework (Alexandru Zamfirache et al., 2023), Hybrid Precoding (HP) (Singh & Shukla, 2022), Binary Anarchic Society Optimization (BASO) (Kilic et al., 2023), Arithmetic optimization algorithm (AOA) (Çelik, 2023). These metaheuristics have been used to solve a variety of optimization issues, proving to be successful and efficient in finding close to ideal solutions in a fair period (Hamad & Rashid, 2023a). In another study, a metaheuristic algorithm was used in the field of fuzzy controllers(Bojan-Dragos et al., 2021). Recently, these behaviors were described in several methods of optimization, an overview of which is shown in Table 1.

**Table 1: A list of Nature-Inspired Algorithms, in which the behaviour of insects or animals inspired the optimisation approach.**

| Algorithms | Nature's inspiration | Author(s) | Year |
|---|---|---|---|
| Particle Swarm Optimization | A swarm of birds, fish, and other animals. | (Kennedy & Eberhart, 1995) | 1995 |
| Ant Colony Optimization | Ants in a colony | (Dorigo & Di Caro, 1999) | 1999 |
| Artificial Bee Colony | Honey bee swarm | (Karaboga & Basturk, 2007) | 2007 |
| Monkey Algorithm | Monkeys | (Zhao & Tang, 2008) | 2008 |
| Firefly Algorithm | Fireflies | (X. S. Yang, 2009) | 2009 |
| Bat Algorithm | Bats | (X. S. Yang, 2010) | 2010 |
| Bacterial Colony Optimization | Escherichia coli | (Niu & Wang, 2012) | 2012 |
| Krill Herd Algorithm | Krills | (Gandomi & Alavi, 2012) | 2012 |
| Cuckoo Search | Cuckoos | (Gandomi et al., 2013) | 2013 |
| Grey Wolf Optimizer | Grey wolves | (Mirjalili et al., 2014) | 2014 |
| Dragonfly algorithm | Dragonflies | (Mirjalili, 2016) | 2016 |
| Donkey and smuggler optimization algorithm | Donkeys | (Shamsaldin et al., 2019) | 2019 |
| Fitness Dependent Optimizer | Bee swarms | (Abdullah & Ahmed, 2019) | 2019 |
| Ant Nesting Algorithm | Ant | (Rashid et al., 2021) | 2021 |
| FOX optimization algorithm | Fox | (Mohammed & Rashid, 2022) | 2022 |

This paper proposes a new algorithm under the name GOOSE algorithm. It is inspired by the swarming behavior of geese during rest and looking for food. The following summarises the main contributions of this paper:

1. The goose standing on one leg was inspired to design the model.

2. Using this technique, a special GOOSE algorithm inspired by nature is created.

3. On the other hand, new algorithms are always welcome as long as they provide equivalent or better functionality. The no-free-lunch idea means that no one optimization algorithm can solve every optimization problem, necessitating the development of new optimization techniques. Consequently, no global method can provide an optimal solution for each optimization issue. For example, there is a strong chance that algorithm "A2" outperforms algorithm "A1" on optimization problem Y if algorithm "A1" performs better than algorithm "A2" for optimization problem X.

4. When we say that method "A1" is asymptotically more efficient than algorithm "A2," we mean that as the quantity of input rises, "A1" outperforms "A2" in terms of time and space complexity. These factors led to the proposal of a novel algorithm in this study, namely GOOSE. The algorithm was tested on various kinds of optimization benchmark test functions and compared to some of the most popular and excellent metaheuristic algorithms like a Genetic Algorithm, Dragonfly Algorithm, Particle Swarm Optimization, Whale Optimization Algorithm, FOX Optimization Algorithm, Salp Swarm Algorithm, and Fitness Dependent Optimization, Butterfly Optimization Algorithm, and Chimp Optimization Algorithm.

5. Using the GOOSE algorithm to optimize four real-world problems in engineering and medicine, Welded beam design, Economic Load Dispatch Problem, pressure vessel design problems, and the pathological IgG fraction in the nervous system.

The rest of the parts of the paper are structured as follows. It starts by outlining the GOOSE algorithm's rationale before debating the special features that highlight its novelty. The Inspirations, Mathematical Framework, and GOOSE Algorithm are described in Section 2. The numerical studies are listed in Section 3, together with the mathematical test functions, findings, explanation, comparison, and statistical test. Finally, the study is concluded in Section 5, which also makes some recommendations for further research.

2. **GOOSE Life, GOOSE Behavior, and GOOSE Algorithm**

The inspiration for the suggested strategy is initially explained in this section. The mathematical framework is then supplied.

**2.1 GOOSE Life**

A goose, multiple geese, is a kind of waterfowl of the Anatidae family. Geese are huge, heavy-bodied birds that are greater than ducks but smaller than swans, and their colour and size may vary depending on the genus. Geese are very sociable creatures. They normally get along with other animals and poultry if they are reared with them. Seeds, nuts, grass, plants, and berries are among the foods consumed by geese. Despite being waterfowl, they spent the majority of their time on land. Gooses always fly in the form of a "V", which gives them an average of 71% more travelling distance. When the leading goose becomes tired, another goose takes over. Goose is very loyal. They mate for a lifetime and are very protective of their spouses and children. When goose loses a relationship mate or their clutch of eggs, they exhibit grieving behaviour. They have deep feelings for the people in their group. If a goose becomes ill or injured, a couple of other geese may leave the flock to take care and safeguard him.

Ancient Rome is the first known instance of a goose being employed for protection. On the other hand, from one extreme to another, geese are deployed to defend the police station along with other facilities in the Chinese province of Xinjiang. According to Mr. Zhang, the police chief, "they are more valuable than dogs in several ways".In addition, In Dumbarton, Scotland, a warehouse where the renowned Ballantine's whiskies were aged was guarded by the goose. From the end of the 1950s, until contemporary cameras were installed, these geese served for roughly 53 years (*Why Geese Are Good Guard Animals | Hello Homestead*, n.d.).

**2.2 GOOSE Behaviour**

In natural environments, every population of goose has one or two guardians who are responsible for guarding while the others forage or rest on a grassy area. The goose standing on one leg (Hamadani et al., 2016). Sometimes, he carries a small stone with his raised leg so that whenever he falls asleep, the stone will burn and the goose will wake up. When the goose hears any strange sound or movement, they make a loud noise to wake up the individuals in the herd and keep them safe. Goose, unlike other birds, respond to what they observe in a manner that is beneficial to humans. Goose has a loud, aggressive sound that is perfect for protection. Goose are highly possessive of their own homes, which they fiercely guard, particularly during the mating and fledging periods. Images of these birds' interactions with other individuals within a group are shown in Figure 1.

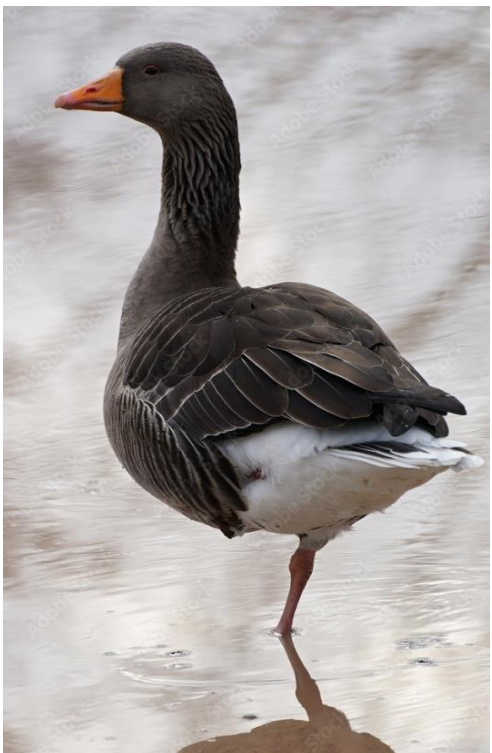 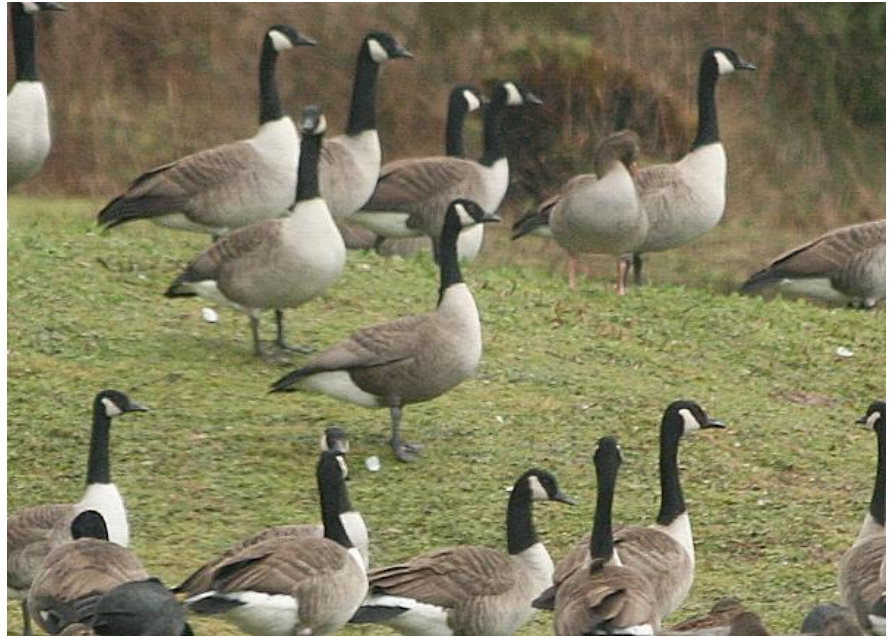

(a) (b)

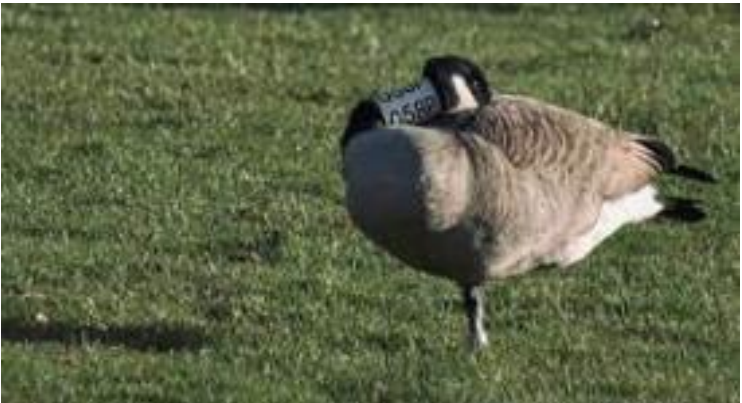 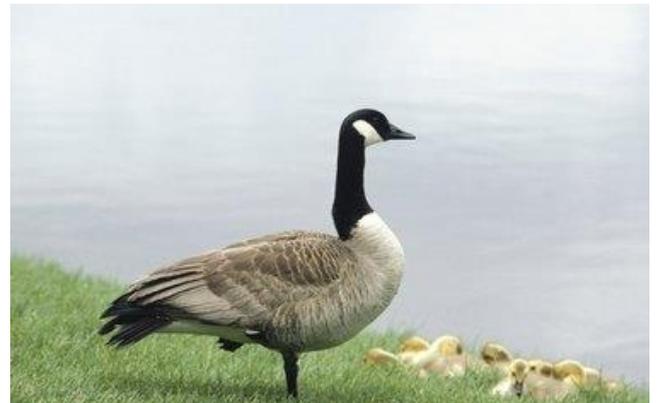

(c) (d)

Fig 1: photograph of behaivir's of GOOSE in the rest: a https://vancouverisland.ctvnews.ca/  b https://lens.google.com/  c https://stock.adobe.com/  d https://paulsbirdingdiary.blogspot.com/

With this behavior, the goose herd has created an attractive protective atmosphere among themselves. This paper based on the behavior of a goose proposed a novel metaheuristics algorithm. The basic techniques are based on the goose trying to stand on one leg and raise a stone on the other leg. The procedures are detailed below:

1. During their repose, geese congregate in groups, with one of them balancing on a single leg.

2. Occasionally, he raises one leg and carries a small stone so that when he falls asleep the stone falls back down, and the goose will awaken.

3. The goose produces a loud honk to alert the others in the herd to keep them safe when they notice any unexpected noises or activities.

In the beginning, geese gather in groups in their shelters and rest areas. Within the population, one of the geese is assigned to guard. He begins to carry out his command by standing and balancing on one leg. Whenever the goose falls asleep, its legs or the stone it is carrying falls to the ground. At this time, the sound of the rock spreads to the other goose in the group. It will be in a state of exploitation upon hearing the sound. As a consequence of this, it takes some time for the other individuals to hear the guardian goose's call. The distance that sound travels may be estimated by multiplying the period by the 343 meters per second that sound travels in the air. Figure 2 shows the guarding behaviour of the goose.

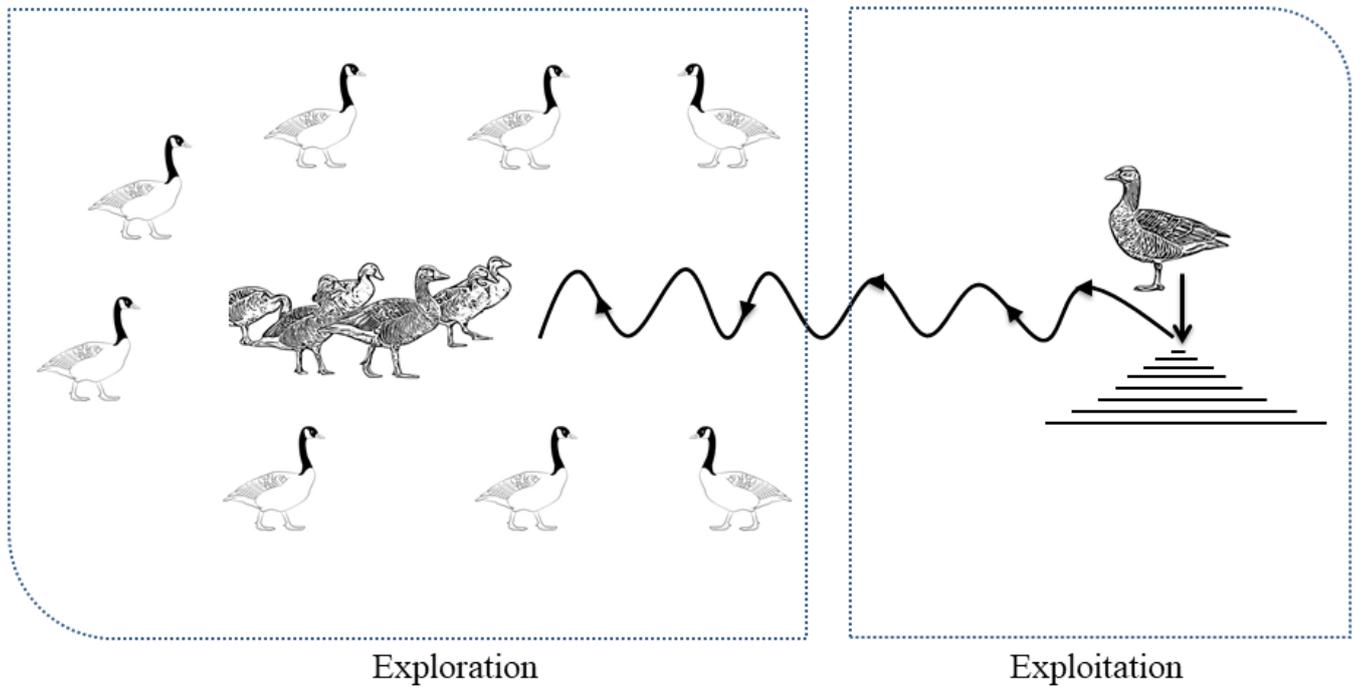

**Fig. 2: Schematic of Goose Behavior, Exploration, and Exploitation**

### 2.3 The GOOSE Algorithm

GOOSE initializes the population, also known as the *X* matrix, first. The location of the goose is an *X*. Return the search agents that go beyond the search space after GOOSE initializes the population. The fitness of each search agent is then determined in each iteration via standardized benchmark functions. The value of fitness of every agent in the search (each line within the *X* grid) is measured and contrasted against the fitness of the remaining agents (other lines) to examine *BestFitness* and *Best Position* (*BestX*). *BestFitness* and *BestX* are operations that compare the fitness of each current row ($fitness_i + 1$), and during iterations, the fitness of the row before it ($fitness_{it}$) is given back.

In the next steps, the exploration and exploitation stages are then balanced by using a condition and a random variable. The value of this variable desires to evenly distribute the phases according to the number of iterations. We give a 50% probability of either exploitation or exploration in GOOSE using the designated random variable, known as "*rnd*". In order, the iterations are split evenly between exploration and exploitation using a conditional phrase. In addition, there are also several other variables introduced, such as *pro*, *rnd*, and *coe*, which randomly find this price. Although the values of variables *pro*, *rnd*, and *coe* were found between 0 and 1. One condition is set to check whether the value of *coe* is less than or equal to 0.17; otherwise, we will equal the value to 0.17. The function of the *pro* variable is to work out which equation works. More than that, you will find the variable weight of the stone carried by the goose with its feet. In the next few sections, exploration and exploitation will be discussed in detail.

#### 2.3.1 Exploitation Phase

The possibility of safeguarding the groups, as described in Section 2.3, is a prerequisite we have for the exploitation phase. We will find the weight of the stone that the goose stores in its feet, which is estimated to be between 5 and 25 grams Through this equation (1), we find the weight of the stone randomly for any iteration. This variable indicates the number of iterations.

$$S\_W_{it} = \in (5,25) \tag{1}$$

Then, in Eq. (2), we should find the time $T\_o\_A\_O_{it}$ needed to reach the earth when the stone falls. It's randomly between 1 and the number of dimensions for each iteration in the loop.

$$T\_o\_A\_O_{it} =\in (1, dim) \quad (2)$$

In Eq. (3), we find the time $T\_o\_A\_S_{it}$ when the object hits the ground and a sound is made and transmitted to the individual goose in the herd.

$$T\_o\_A\_S_{it} =\in (1, dim) \quad (3)$$

In the next equation, discover the total time required for the sound to propagate and reach the individual goose in the flock throughout the iterations. As shown in Eq. (4), the total amount of time is divided by the dimensions. To obtain the average time required, we divide the total time by 2. Eq. (5) explains the steps.

$$T\_T = \frac{\sum(T\_o\_A\_O_{it} + T\_o\_A\_S_{it})}{dim} \quad (4)$$

$$T\_A = \frac{T\_T}{2} \quad (5)$$

As we discussed in the previous sections, there is a random variable *rnd* responsible for the distribution of the exploitation and exploration phases. The value of variable *pro* is randomly selected from the range [0, 1]. Consider the value of variable *pro* is greater than 0.2 and $S\_W_{it}$ greater than or equal to 12. In Eq. (6), $T\_o\_A\_O_{it}$ is multiplied by the square root of the $S\_W_{it}$, after that multiplied by the object's acceleration (g) at 9.81 meters per square second, $M/S^2$. These equations should be used to protect and awaken the individual in the group.

$$F\_F\_S = T\_o\_A\_O_{it} * \sqrt[2]{S\_W_{it}} * g \quad (6)$$

In Eq.(7), to find the distance of sound travel $D\_S\_T_{it}$, it must be the speed of sound $S\_S$ in the air multiplied by the time of sound travel $T\_o\_A\_S_{it}$. The speed of sound is 343.2 meters per second in the air. Figure 3 explains the distance sound travels (Igel, 2012).

$$D\_S\_T_{it} = S\_S * T\_o\_A\_S_{it} \quad (7)$$

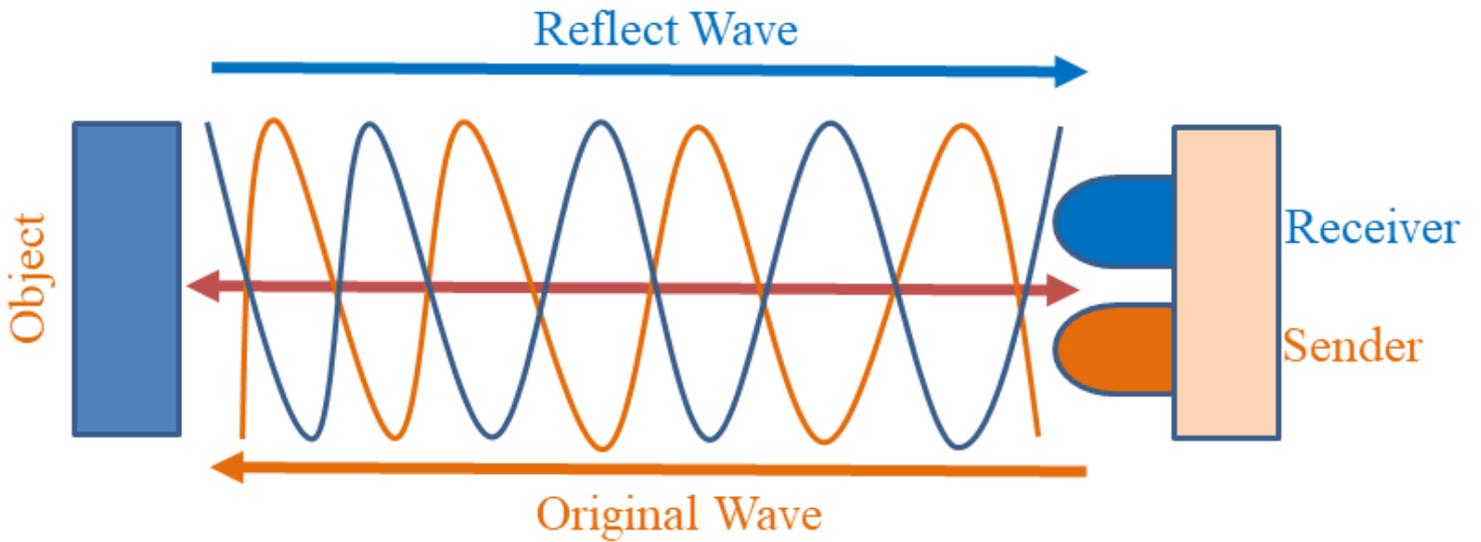

Fig. 3: Distance of sound travel in the air

In this step, we find $D\_G_{it}$ the distance between the guard goose and another goose that is resting or feeding. In Eq. (8), we use the distance of sound travel $D\_S\_T_{it}$ multiplied by 1/2 or 0.5 because we only need the time for the sound to travel and not the time for the sound to return.

$$D\_G_{it} = 0.5 * D\_S\_T_{it} \tag{8}$$

To resolve a new X in the population. In other words, to wake up the individual in the flocks, we must find a *BestX_{it}*, as demonstrated in Eq.(9). This equation is composed of the $Current\_position\ or\ (Xt)$ add by the speed of the falling object $\_F\_S$, after that added to the distance of the Goose $D\_G_{it}$ and multiplied by the average of the time squared $T\_A$, to determine of a new X in the population.

$$X_{(it+1)} = Xt + F\_F\_S + D\_G_{it} * T\_A^{\wedge 2} \tag{9}$$

On the contrary, if both variables are the weight of the stone $S\_W_{it}$ and *pro*, one after the other less than 12 and less than or equal to 0.2, find the new X as shown in Eq. (11) below. To obtain the speed of a falling object $F\_F\_S$, multiply the time $T\_o\_A\_O_{it}$ taken to arrive at the object by the weight of the stone $S\_W_{it}$ multiplied by gravity. In addition, to determine the distance of sound travel $D\_S\_T_{it}$ and the distance of the goose $D\_G_{it}$, we dramatically used the previous equations (7) and (8).

$$F\_F\_S = T\_o\_A\_O_{it} * S\_W_{it} * g \tag{10}$$

In the other way, we find a new X in the new mathematical equation. In Eq.(3.11), the $Current\_Position\ or\ (Xt)$ added to the speed of the falling object, distance of the goose, average time, and *coe*, are multiplied by each other in succession.

$$X_{(it+1)} = Xt + F\_F\_S * D\_G_{it} * T\_A^{\wedge 2} * Coe \tag{11}$$

In the exploitation phase, we used two equations to discover a new X, for instance, Eq. (9) and Eq. (11). These values of variables *pro* and $S\_W_{it}$ determined which equation was performed.

### 2.3.2 Exploration Phase

In this phase, the goose awakens randomly following the best position that has been discovered so far to regulate the random wake-up or safeguard the individual. In case the goose is not carrying stones with its feet, but randomly individuals in the flock wake up. As soon as one of the geese wakes up, they start screaming to protect all the individuals in the flock. As is obvious from what we have already mentioned in the previous sections if the value of the variable *rnd* is smaller than 0.5, then these equations are applied, such as Eq.(3) and Eq.(4). Coupled with checking that the value of minimum time $M\_T$ is greater than the total time $T\_T$, the minimum time is assigned equal to total time.

The value of variable *alpha* ranges from 2 to 0. This value is dramatically decreased with each iteration in the loop. Eq. (12) is used to improve the result of a new X in the search space.

$$alpha = \left(2 - \left(\frac{loop}{\frac{Max\_It}{2}}\right)\right) \tag{12}$$

Where *Max_It* is the number of iterations that can be made. To shift the search phase in the direction of the answer that is most likely to be the optimal solution, computing the two parameters $M\_T$ and alpha is crucial.

Make sure that the goose stochastically explores the other individuals in the search space by using *randn(1, dim)*. Nevertheless, $M\_T$ and *alpha* variables are both utilized to improve the searchability of GOOSE. In Eq. (13), the minimum of time and alpha are multiplied by a random number, and then added to the best position in the search space.

$$X_{(it+1)} = randn(1, dim) * (M_T * alpha) + Xt \tag{13}$$

Where *dim* is the number of problem dimensions and *Best_pos* is the BestX or best position we found in the search area.

In summary, the goose algorithm randomly starts generating populations. Then, in the first iteration, it reviews the values of the population in the herd to restore the values outside the boundary. Also, implement object functions to determine the best score and the best position within the search boundary. To control the exploitation and exploration phases, we used a random variable *rnd* with randomly selected values. If the value of *rnd* is greater than or equal to 0.5, then the explore phase is activated. Within the limits of this condition, we have two other random variables, such as the *pro* and the weight of the stone *S_W*. If *pro* is greater than 0.2 and the weight of the stone is greater than or equal to 12, equations (1), (2), (3), (4), (5), (6), (7), (8) and (9) apply. On the other hand, if the *pro* is smaller than or equal to 0.2 and the weight of the stone *S_W* is smaller than 12, equations

(1), (2), (3), (4), (5), (7), (8), (10) and (11) apply. In a scenario where *rnd* is less than 0.5, exploration is initiated. To illustrate this, we found these variables *T_o_A_S*, *T_T,* and *alpha*. In this situation, in order equations (3),(4),(12), and (13) are applied.

Furthermore, in this algorithm, three methods are adapted to find a new X in the search space. The exploitation phase used two equations for instance Eq.(9) and Eq.(11). But, in the exporation phase only one equation, Eq.(13) to detect a new X.

### 2.3.3 Flowchart and Pseudo-Code of GOOSE

In this section, we have explained the flowchart and pseudocode of the GOOSE algorithm as shown in Figure (4) and Algorithm (1).

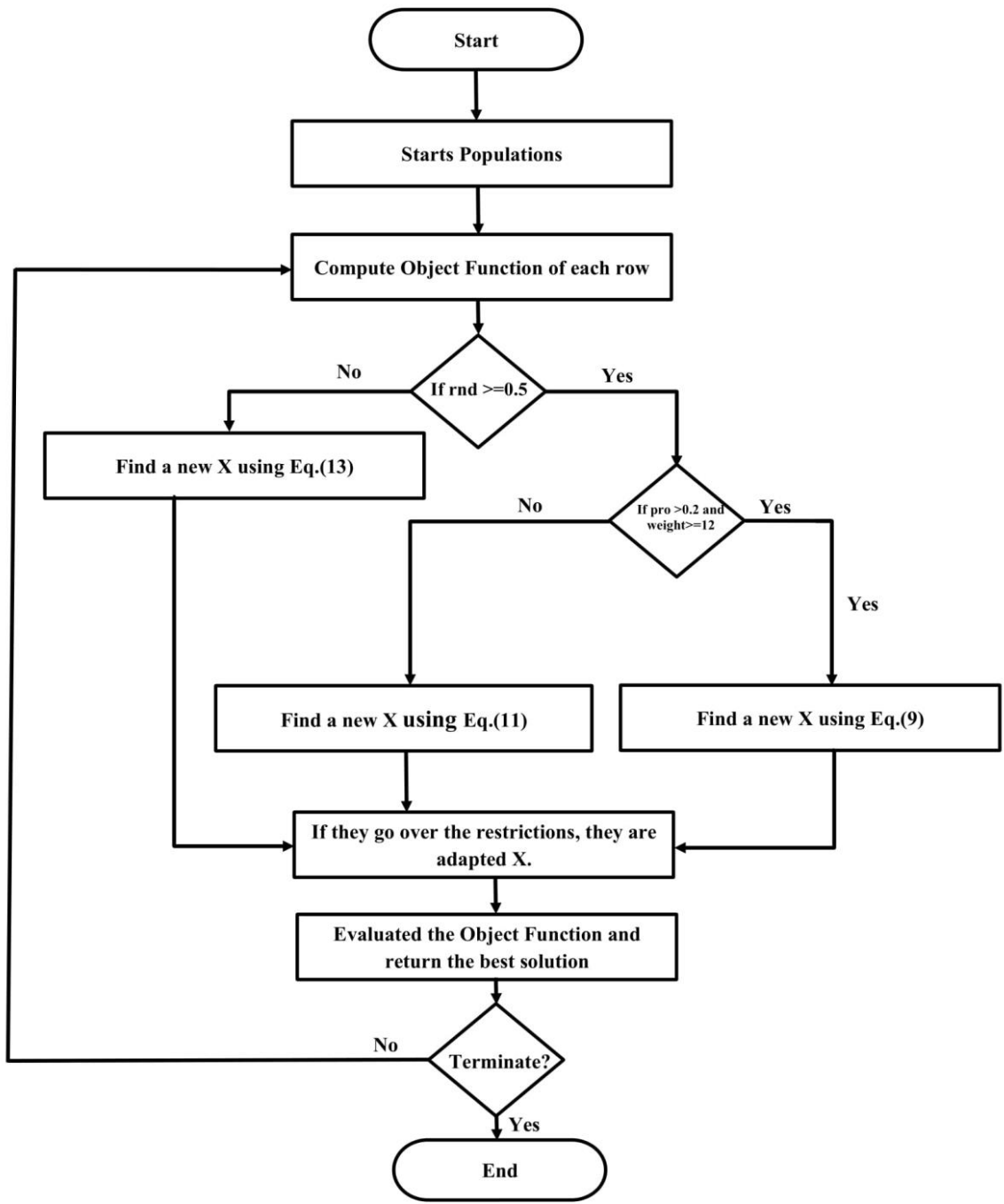

**Fig. 4: Flowchart of GOOSE Algorithm**

**Algorithm 1 GOOSE Algorithm**

1: Initialize the goose population Xi (i=1,2,……,n)
2: While loop<Max_It
3: Generate **D_S_T, T_o_A_S, BestX, D_G, M_T, alpha, BestFitness**
4: Calculate the object function of each search agent
5: Select BestX and BestFitness among the goose population (X) in each iteration
6: **If$_1$** for checking each agent
7: Update the current position of the search agent
8: **Endif$_1$**
9: **For**
10: Calculate rnd, pro and coe
11: Find S_W using Eq.(1)
12: Calculate time randomly, using Eq.(2) and Eq.(3)
13: Calculate T_T and T_A using Eq.(4) and Eq. 5)
14: **If2** rnd>=0.5
15: **If3** pro>0.2 and W_S>=12
16: Find F_F_S using Eq.(6)
17: Calculate D_S_T using Eq,(7)
18: Calculate D_G using Eq. (8)
19: Find $X_{(it+1)}$ using Eq. (9)
20: **Else**
21: Find F_F_S using Eq.(10)
22: Calculate D_S_T using Eq. (7)
23: Calculate D_G using Eq. (8)
24: Find $X_{(it+1)}$ using Eq. (11)
25: **EndIf3**
26: **Else**
27: Find alpha using Eq.(12)
28: Explore $X_{(it+1)}$ using Eq. (13)
29: **EndIf2**
30: If they go over the restrictions, they are adapted X.
31: Evaluate search agents by their fitness
32: Update BestX
33: **EndFor**
34: *Loop=Loop*+1
35: **End while**
36: Return BestX & BestFitness
37: Initialize the goose population Xi (i=1,2,……,n)
38: While loop<Max_It

## 3. Implementation and Discussion

To ensure the proper performance of the proposed algorithm, it should be tested using different benchmark functions. To ensure the proper performance of the proposed algorithm, it should be tested using different benchmark functions. To evaluate the performance of the Goose algorithm, we tested different benchmark functions on the algorithm, such as classical benchmarks and recent benchmark functions.

Every benchmark function was subjected to 30 runs of the GOOSE algorithm. Tables 2-4 provide the results of the statistical analysis (mean and standard deviation). Additionally, the suggested algorithms' outcomes are contrasted with seven Swarm Intelligence-based techniques [SIs]: FDO, DA, PSO, BOA, WOA, SSA, and ChOA. In addition, the GOOSE algorithm is compared with GA as an Evolutionary Algorithm[EAs] and FOX as Nature-Inspired Optimization[NIOs]. The significance of the findings is then assessed by statistical comparison of these outcomes to one another. Therefore, the Welded beam design, Economic Load Dispatch Problem, and Pressure Vessel Design Problem three renowned real-world engineering challenges, and the Pathological IgG Fraction in the Nervous System are solved using the GOOSE to make sure that it performs well in solving real-world applications.

### 3.1 Standard benchmark functions

Several widely used benchmarks or test functions may be used to evaluate the validity, effectiveness, and dependability of optimization methods. In the literature on metaheuristics, benchmark numerical test functions are often employed as instruments for assessing performance. It is also believed that metaheuristic algorithms that perform these functions may resolve challenging optimisation issues in the real world. The benchmark functions, which may be categorised into four types unimodal, multimodal, fixed-dimension multimodal, and composite functions, are generally minimization functions. Even though they are straightforward, we have selected these test functions so that we may contrast our findings with those of the most recent meta-heuristics. These benchmark functions are presented in Appendix A Tables A1-A6, where *Dim* denotes the function's dimension, *Range* represents the search space's boundary, and *fmin* is the optimum value. These benchmark functions represent the variations of the classical functions that have been moved, rotated, enlarged, and combined to give the highest level of sophistication (Suganthan et al., 2005).

As a result, 19 popular standard benchmark functions with a variety of features are chosen to evaluate the algorithm's performance. Unimodal, multimodal, and composited test functions are included in the sets. There is just one optimal value for the unimodal test functions. They are used to evaluate exploitation potential. They enable concentrating more on the algorithm's convergence rate as opposed to the outcomes. The six unimodal test functions F1, F2, F3, F4, F5, and F7—that were chosen to test the GOOSE algorithm are shown in Table A1 in Appendix A.

The multimodal test functions have several optimal locations, and as the number of issue dimensions rises, so does the number of local optimum locations. They are used to assess an algorithm's capacity for exploration, which may help it steer clear of local optima. The six multimodal test functions—F8, F9, F10, F11, F12, and F13—selected to test the GOOSE algorithm are listed in Table A2 in Appendix A. The composite test functions are condensed, shifted, rotated, and biased counterparts of the other test functions, as their name implies. They provide several local optima and a wide variety of forms. They make it possible to gauge the algorithm's balance between exploitation and exploration. The six composite test functions F14, F15, F16, F17, F18, and F9 chosen to test the method are shown in Table A3 in Appendix A. Our suggested GOOSE algorithm is put up against two sets of rival algorithms—each with a distinct set of parameter settings—for verification and analysis.

### 3.1.1 Standard Benchmark Test Functions (FDO, DA, PSO, and GA)

For comparison with GOOSE on the 19 chosen standard benchmark functions, the common FDO, DA, PSO, and GA algorithms are chosen as reference algorithms in the first set. The algorithms GA and PSO are among the oldest, most well-known, and most effective in the literature(Katoch et al., 2021), whereas FDO and DA are more modern, promising algorithms with many successful applications. The test results of the FDO, DA, PSO, and GA algorithms on the 19 chosen standard benchmark functions are included in the reference (Abdullah & Ahmed, 2019), along with a detailed description of the parameter settings. The population size is set to 30, and the dimension of the benchmark functions is set to 10 for the common parameter sets in each scenario. The halting condition is set at 500 iterations, which is the maximum number possible. After 30 iterations of the program, the average and standard deviation are determined. The test results for the GOOSE, DA, PSO, and GA algorithms on the 19 common benchmark functions are shown in Table 2.

As indicated in Table 2, each algorithm test function derived from the common benchmark functions is minimized toward 0.0. When GOOSE's test results were compared to those of the other algorithms in the table, GOOSE surpassed the most well-known algorithms: On eight test scenarios, namely F7, F9, F11, and F15–F19, FDO, DA, PSO, and GA all outperformed GOOSE, except F10, F12, and F14. The outcomes of F10, F12, and F14 were, however, not subpar—only inferior to those of the other algorithms. On the remaining benchmark functions, the algorithm delivered results that were comparable to those of the others.

The findings of Table 2 show that the composite test functions F13 through F19 are acceptable for testing an algorithm's avoidance of local minima. Except for F13 and F14, which came in third place with the outperformance of the DA and PSO algorithms and second place with the outperformance of the FDO algorithm, the GOOSE algorithm outscored all the GA, FDO, PSO, and DA algorithms on all of these test functions. This leads to the conclusion that GOOSE is very good at avoiding local minima, which balances the scope of exploitation and exploration. Figures 4 through 6 show the benchmark functions' 2D versions.

**Table 2: GOOSE, FDO, DA, PSO, and GA test results on the standard benchmark functions** (Abdullah & Ahmed, 2019).

| Test Function | GOOSE | | FDO | | DA | | PSO | | GA | |
|---|---|---|---|---|---|---|---|---|---|---|
| | **Avg** | **Stdv** | **Avg** | **Stdv** | **Avg** | **Stdv** | **Avg** | **Stdv** | **Avg** | **Stdv** |
| **F1** | 1.15E-05 | 1.84E-05 | **7.47E-22** | 7.26E-19 | 2.85E-18 | 7.16E-18 | 4.2E-18 | 1.31E-17 | 7.49E+02 | 3.25E+02 |
| F2 | 1.16E-02 | 7.93E-03 | **9.388E-07** | 6.91E-06 | 1.49E-05 | 3.76E-05 | 0.003154 | 0.009811 | 5.971358 | 1.533102 |
| F3 | 0.0011 | 1.50E-03 | **8.552E-08** | 4.40E-06 | 1.29E-06 | 2.1E-06 | 0.001891 | 0.003311 | 1949.003 | 994.2733 |
| F4 | 1.00E-03 | 8.19E-04 | **6.688E-05** | 2.49E-03 | 0.000988 | 0.002776 | 0.001748 | 0.002515 | 21.16304 | 2.605406 |
| F5 | 2.88E+01 | 2.19E-02 | 23.501 | 5.98E+01 | **7.600558** | 6.786473 | 63.45331 | 80.12726 | 133307.1 | 85007.62 |
| F6 | 0.0099 | 3.32E-03 | **1.422E-19** | 4.75E-18 | 4.17E-16 | 1.32E-15 | 4.36E-17 | 1.38E-16 | 563.8889 | 229.6997 |
| F7 | **5.70E-03** | 3.82E-03 | 0.544401 | 3.15E-01 | 0.010293 | 0.004691 | 0.005973 | 0.003583 | 0.166872 | 0.072571 |
| F8 | -7187.6 | 6.59E+02 | -2285207 | 2.07E+05 | -2857.58 | 383.6466 | **-7.1E+11** | 1.2E+12 | -3407.25 | 164.4776 |
| F9 | **0.0038** | 5.31E-03 | 14.56544 | 5.20E+00 | 16.01883 | 9.479113 | 10.44724 | 7.879807 | 25.51886 | 6.66936 |
| F10 | 0.002 | 2.07E-03 | **3.996E-16** | 6.38E-16 | 0.23103 | 0.487053 | 0.280137 | 0.601817 | 9.498785 | 1.271393 |
| F11 | **6.67E-07** | 9.68E-07 | 0.568776 | 1.04E-01 | 0.193354 | 0.073495 | 0.083463 | 0.035067 | 7.719959 | 3.62607 |
| F12 | 0.00026 | 1.18E-04 | 19.83835 | 2.64E+01 | 0.031101 | 0.098349 | **8.57E-11** | 2.71E-10 | 1858.502 | 5820.215 |
| F13 | 0.0079 | 6.85E-03 | 10.2783 | 7.42E+00 | **0.002197** | 0.004633 | 0.002197 | 0.004633 | 68047.23 | 87736.76 |
| F14 | 9.9012 | 3.90E+00 | **3.787E-08** | 6.32E-07 | 103.742 | 91.24364 | 150 | 135.4006 | 130.0991 | 21.32037 |
| F15 | **0.000315** | 1.38E-05 | 0.0015202 | 1.24E-03 | 193.0171 | 80.6332 | 188.1951 | 157.2834 | 116.0554 | 19.19351 |
| F16 | **-1.0316** | 6.66E-16 | 0.006375 | 1.06E-02 | 458.2962 | 165.3724 | 263.0948 | 187.1352 | 383.9184 | 36.60532 |
| F17 | 0.3979 | **1.67E-16** | 23.82013 | 2.15E-01 | 596.6629 | 171.0631 | 466.5429 | 180.9493 | 503.0485 | 35.79406 |
| F18 | 3 | 0 | 222.9682 | 9.96E-06 | 229.9515 | 184.6095 | 136.1759 | 160.0119 | 118.438 | 51.00183 |
| F19 | **-3.8628** | 3.11E-15 | 22.7801 | 1.04E-02 | 679.588 | 199.4014 | 741.6341 | 206.7296 | 544.1018 | 13.30161 |

\* Note: The bolded values show that the algorithm produced the best results when compared to the other algorithms.

### 3.1.2 Classical Benchmark Test Functions (GOOSE, FDO, FOX, BOA, WOA, DA, and ChOA)

In the second set, the GOOSE algorithm and four other algorithms (FDO, FOX, BOA, and WOA) are chosen as reference algorithms for comparing to GOOSES on five chosen standard benchmark functions (F1, F5, F8, F9, and F11). The test results of the GOOSE, FDO, FOX, BOA, and WOA algorithms on the five standard benchmark functions used in this study (Mohammed & Rashid, 2022). The parameter settings are shown in detail in Appendix A Table A6.

In all situations, the population size is set to 30, and the dimension of the benchmark functions is set to 30. The halting criterion is set at 500 iterations as the maximum number of iterations. It should be noted that the functions are utilized without shift, and the range is decreased to [5.12,-5.12]. The method is run 30 times to get the average and standard deviation. Table 3 shows the performance of the GOOSE, FDO, FOX, BOA, and WOA algorithms on the five standard benchmark functions.

As demonstrated in Table 3, each test function of the method from the typical benchmark functions is outstandingly minimized towards 0.0. When the test results of GOOSE were compared to the other algorithms in the table, GOOSE beat these algorithms: FDO, FOX, BOA, and WOA on one test instance, namely F8. The GOOSE algorithm performed comparably to the others on the remaining benchmark functions.

Table 3: GOOSE, FDO, FOX, BOA, WOA, DA, and ChOA test results on the Five Classical benchmark functions(Mohammed & Rashid, 2022).

| Test Function | GOOSE | | FDO | | FOX | | BOA | | WOA | | DA | | ChOA | |
|---|---|---|---|---|---|---|---|---|---|---|---|---|---|---|
| | Avg | Stdv | Avg | Stdv | Avg | Stdv | Avg | Stdv | Avg | Stdv | Avg | Stdv | Avg | Stdv |
| F1 | 0.089355 | 0.32601609 | 7.47E-21 | 7.26E-19 | 0 | 0 | 1.01E-11 | 1.66E-12 | 1.41E-30 | 4.91E-30 | 2.85E-18 | 7.16E-18 | 6.86E-49 | 3E-08 |
| F5 | 112.5549 | 111.865891 | 23.501 | 59.7883 | 38.4337 | 0.082471 | 8.9355 | 0.0215 | **0.072581** | 0.39747 | 7.6005 | 6.7864 | 27.1546 | 0.001624 |
| F8 | **-7208.68** | 718.060211 | -2285207 | 206684 | 0 | 0 | NA | NA | -5080.76 | 695.7968 | -2857 | 383.64 | 5.68E-14 | 0.001203 |
| F9 | 141.4872 | 28.4285865 | 14.5654 | 5.2022 | 0 | 0 | 28.68 | 20.178 | 0 | 0 | 16.0188 | 9.4791 | 0 | 0 |
| F11 | 158.6918 | 195.756062 | **0.5687** | 0.1042 | -6097.8 | 387.2942 | 1.35E-13 | 6.27E-14 | 0.000289 | 0.001586 | 0.1933 | 0.0734 | -3628.802 | 5.1249 |

### 3.2. CEC-C06 2019 Benchmark Test Functions

A set of ten contemporary CEC benchmark functions are utilized in addition to the traditional benchmark functions to evaluate the GOOSE algorithm further. The findings are compared to those of the three other notable metaheuristic algorithms, DA, WOA, and SSA. These "100-digit challenge" test functions, which are created for benchmarking single-objective optimization problems, are meant to be utilized in yearly optimization contests (K. V. Price, N. H. Awad, M. Z. Ali, 2018). The CEC-06 2019 test functions for GOOSE algorithm benchmarking are included in Table A5 in Appendix A.

All test functions for CEC-06 2019 are scalable; however, only test functions CEC04 to CEC10 are capable of being rotated or relocated, unlike CEC01 to CEC03. The test of GOOSE is run using the default test function settings that the CEC benchmark creator supplied. Table A5 in Appendix A shows that function CEC01 is set as a 9-dimensional minimization problem in the boundary range [-8192, 8192], function CEC02 is set as 16 dimensional in the boundary range [-16,384, 16,384], function CEC03 is set as 18 dimensional in the boundary range [-4,4], and the remaining functions, from CEC04 to CEC10, are set as a 10-dimensional minimization problem in the boundary range [-100,100]. For greater ease, the global optimum of all CEC functions converged to 1.0.

The test results of three other contemporary optimization algorithms—SSA, DA, and WOA—taken from Abdullah and Rashid (Abdullah & Ahmed, 2019) are compared to those of the GOOSE algorithm. Concerning standard parameter settings, the same ones as those previously used (Abdullah & Ahmed, 2019) are applied, with 500 iterations and 30 agents. The average and standard deviation for each test function is calculated after the method has been performed 30 times. The test results for the CEC-C06 2019 test functions for GOOSE, DA, WOA, and SSA are shown in Table 4.

It is clear from Table 4 that each test function for the GOOSE algorithm on CEC functions is minimized in the direction of one. On all other test scenarios, GOOSE fared better than all the other algorithms. This is just another demonstration of the superior performance and effectiveness of the GOOSE algorithm. It is important to note that on the CEC03 function, the WOA algorithm yields the same outcome as GOOSE. However, WOA's standard deviation for the CEC03 function is 0.0, which suggests there is no need for improvement since WOA consistently produces the same result.

Table 4: GOOSE, DA, WOA, and SSA test results on the Ten benchmark functions(Abdullah & Ahmed, 2019).

| Test Function | GOOSE | | DA | | WOA | | SSA | |
|---|---|---|---|---|---|---|---|---|
| | Avg | Stdv | Avg | Stdv | Avg | Stdv | Avg | Stdv |
| CEC01 | 1.8823E+12 | 2.13E+12 | 54300000000 | 6.69E+10 | 4.11E+10 | 5.42E+10 | **6.05E+09** | 4.75E+09 |
| CEC02 | 6013.8 | 8365.468 | 78.0368 | 87.7888 | **17.3495** | 0.0045 | 18.3434 | 0.0005 |
| CEC03 | **13.7024** | 7.11E-15 | 13.7026 | 0.0007 | **13.7024** | 0 | 13.7025 | 0.0003 |
| CEC04 | 1710.6 | 969.4561 | 344.3561 | 414.0982 | 394.6754 | 248.5627 | **41.6936** | 22.2191 |
| CEC05 | 6.0916 | 1.652374 | 2.5572 | 0.3245 | 2.7342 | 0.2917 | **2.2084** | 0.1064 |
| CEC06 | **4.7857** | 0.909049 | 9.8955 | 1.6404 | 10.7085 | 1.0325 | 6.0798 | 1.4873 |
| CEC07 | **274.3512** | 238.7282 | 578.9531 | 329.3983 | 490.6843 | 194.8318 | 410.3964 | 290.5562 |
| CEC08 | **5.5691** | 0.560903 | 6.8734 | 0.5015 | 6.909 | 0.4269 | 6.371723 | 0.5862 |
| CEC09 | 3.807 | 0.356037 | 6.0467 | 2.871 | 5.9371 | 1.6566 | **3.6704** | 0.2362 |
| CEC10 | **20.9835** | 0.029484 | 21.2604 | 0.1715 | 21.2761 | 0.1111 | 21.04 | 0.078 |

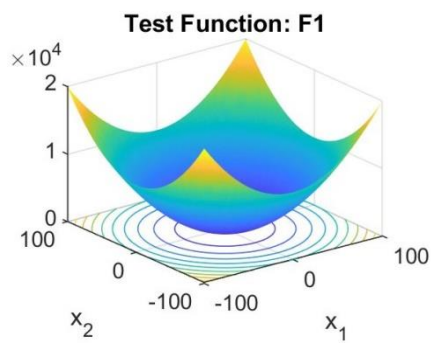
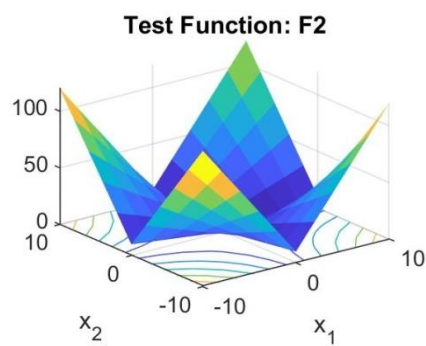
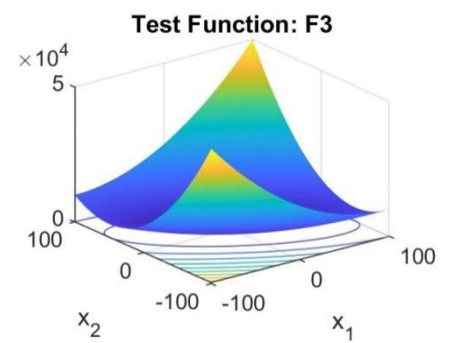

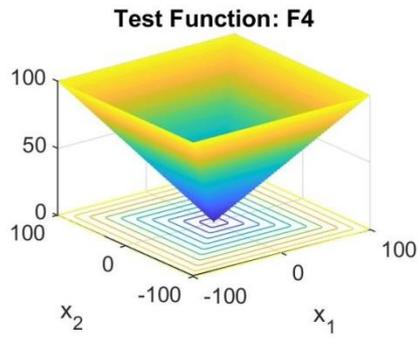
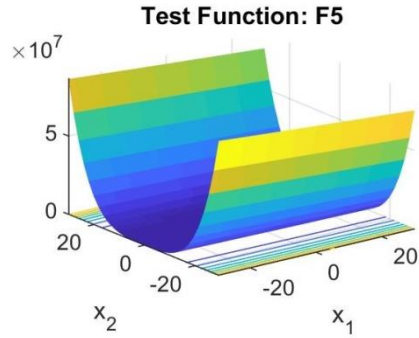
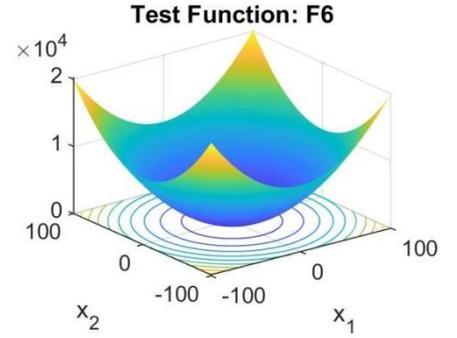
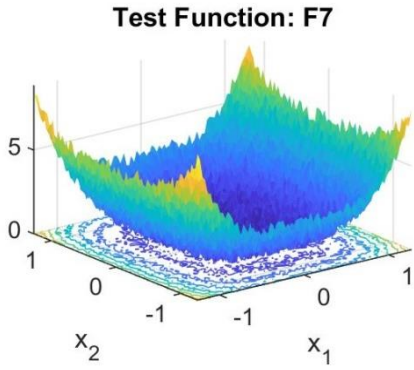

Fig. 5: 2-D versions of unimodal benchmark functions

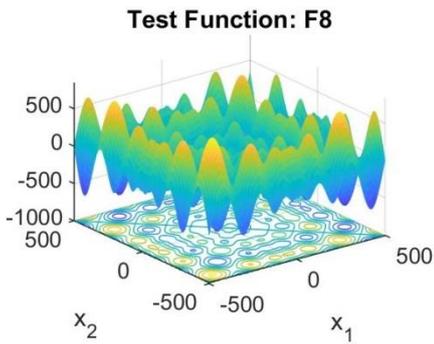
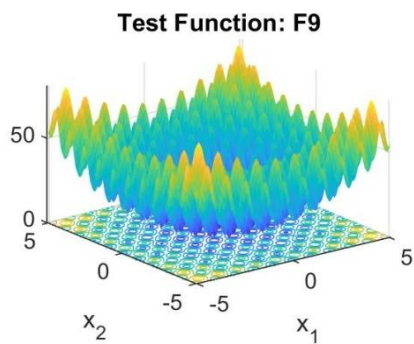
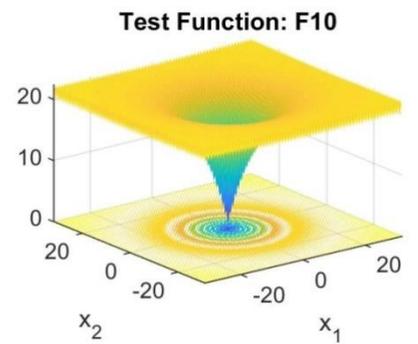
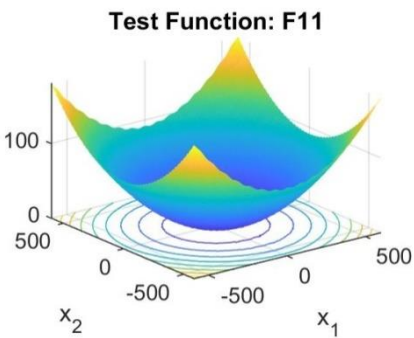
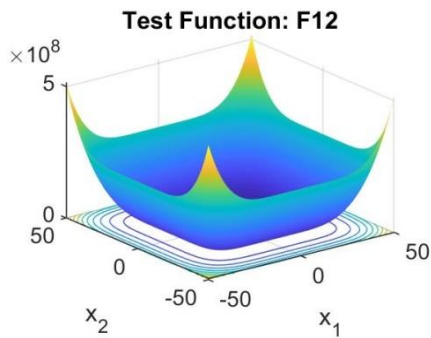
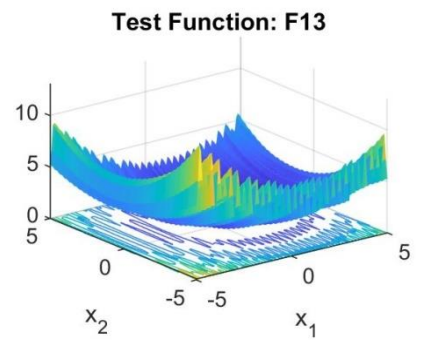

Fig. 6: 2-D versions of multimodal benchmark functions

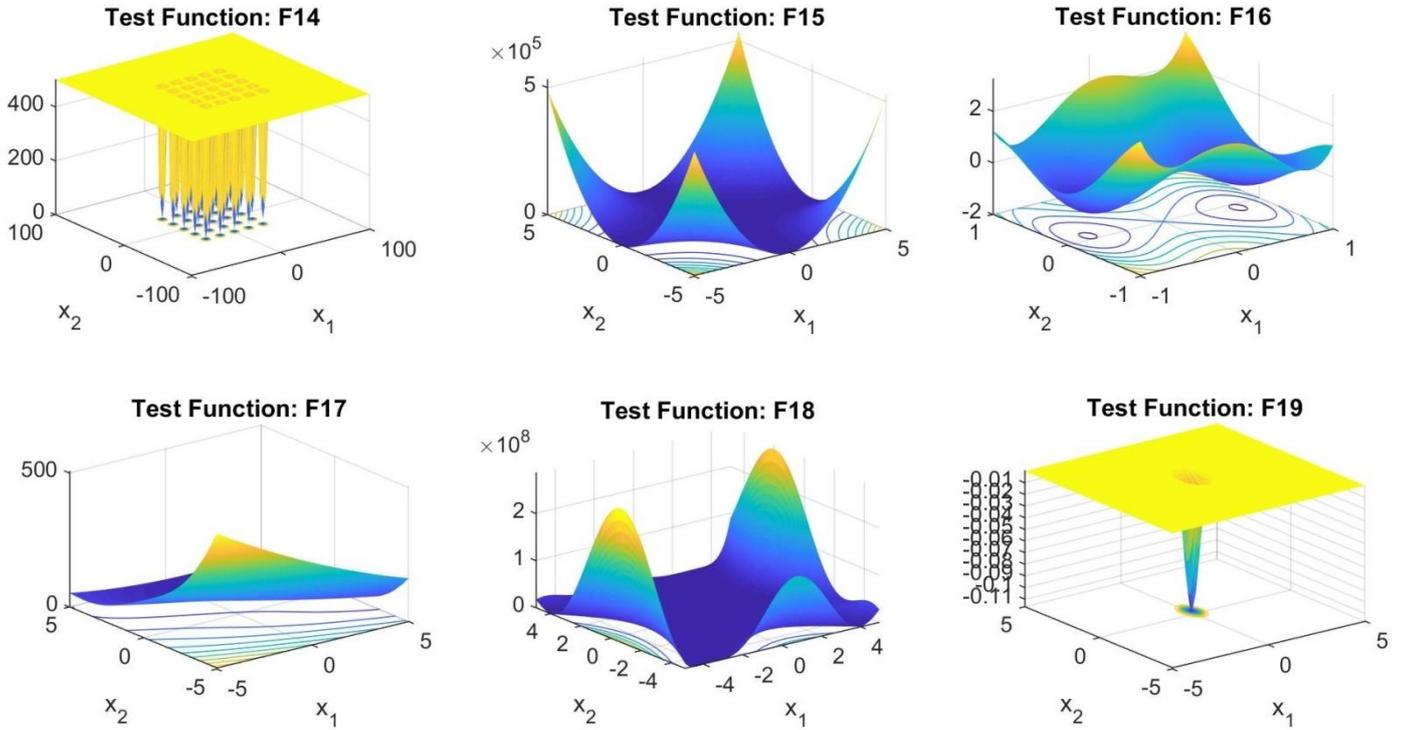

**Fig. 7: 2-D versions of fixed-dimension multimodal benchmark functions**

### 3.3 Comparative Study

The effectiveness of algorithms may be compared using a variety of metrics and methods. Given the significance of achieving optimality in optimization, this section compares the global average best solutions of the GOOSE, GA, FDO, PSO, and DA algorithms with those of the GOOSE, FDO, FOX, BOA, and WOA algorithms and the GOOSE, DA, WOA, and SSA algorithms on the most common benchmark functions used to test the GOOSE algorithm. On the 19 common benchmark functions, Tables 5 and 6 show the comparative study of the GOOSE algorithm and the rankings of GOOSE, DA, PSO, FDO, and GA algorithms. Table 7 shows the total number of first, second, third, fourth, and fifth ranks for the algorithms. Table 8 shows how the GOOSE, FDO, FOX, DA, BOA, WOA, and ChOA algorithms rank on the five traditional benchmark functions, and Table 9 shows how the GOOSE, SSA, DA, and WOA algorithms rank on the ten modern standard benchmark functions. The entire number of the algorithms' first, second, third, and fourth ranks is also shown in Table 10.

Tables 5, 6, and 7 show that, in contrast to the well-known FDO, DA, PSO, and GA algorithms, the GOOSE algorithm has the highest first-ranking number of eight and the lowest fifth ranking of just zero. Furthermore, Tables 9 and 10 show that GOOSE once again proves to be efficient by outperforming SSA, DA, and WOA algorithms to get the highest first rank on the ten current standard benchmark functions. Moreover, Table 8 shows that the GOOSE algorithm out of 5 functions in the F8 function obtained the first rank on the five classical standard benchmark functions in comparison to FDO, FOX, DA, BOA, WOA, and ChOA algorithms.

To provide a more thorough assessment of the algorithm, GOOSE is compared to the FDO, DA, PSO, and GA algorithms by type of benchmark function and overall about the standard benchmark functions. Tables 5 and 11 show the GOOSE rankings for several benchmark functions, both individually and collectively. Some mechanisms based on the techniques already used by other researchers were utilized to test and verify the findings obtained after implementing the suggested algorithm. The GOOSE algorithm's performance was assessed, and the test result showed that it performed pretty well, earning 2.158 rankings among 19 benchmark functions. Throughout the testing process, the effectiveness of the suggested algorithm was assessed by resolving each of the 19 benchmark functions. Different levels of ranking were observed for various types of problems, for instance, GOOSE ranked 3.143 among the other algorithms for Unimodal functions, which are F1–F7. This demonstrates GOOSE is very adept at seeking out novel solutions, as seen by its rankings of 2 in Multimodal benchmark functions, which include F8-F13, and 1.167 in Composite functions, which include F14-F19. This demonstrates the algorithm's ability to avoid local minima as they thoroughly explore promising locations inside the design space and use the optimal solution. It is important to remember that no method can provide the optimum results for every optimization task. On certain tasks, some algorithms will do far better than others while others will fall short(Cortés-Toro et al., 2018).

**Table 5: A comparison study of the GOOSE Algorithm**

| Test Function | 1st | 2nd | 3rd | 4th | 5th | Rank | Subtotal |
|---|---|---|---|---|---|---|---|
| F1 | FDO | DA | PSO | GOOSE | GA | 4 | |
| F2 | FDO | DA | PSO | GOOSE | GA | 4 | |
| F3 | FDO | DA | GOOSE | PSO | GA | 3 | |
| F4 | FDO | DA | GOOSE | PSO | GA | 3 | 22 |
| F5 | DA | FDO | GOOSE | PSO | GA | 3 | |
| F6 | FDO | PSO | DA | GOOSE | GA | 4 | |
| F7 | GOOSE | PSO | DA | GA | FDO | 1 | |
| F8 | PSO | FDO | GOOSE | GA | DA | 3 | |
| F9 | GOOSE | PSO | FDO | DA | GA | 1 | |
| F10 | FDO | GOOSE | DA | PSO | GA | 2 | 12 |
| F11 | GOOSE | PSO | DA | FDO | GA | 1 | |
| F12 | PSO | GOOSE | DA | FDO | GA | 2 | |
| F13 | DA | PSO | GOOSE | FDO | GA | 3 | |
| F14 | FDO | GOOSE | DA | GA | PSO | 2 | |
| F15 | GOOSE | FDO | GA | PSO | DA | 1 | |
| F16 | GOOSE | FDO | PSO | GA | DA | 1 | 7 |
| F17 | GOOSE | FDO | PSO | GA | DA | 1 | |
| F18 | GOOSE | GA | PSO | FDO | DA | 1 | |
| F19 | GOOSE | FDO | GA | DA | PSO | 1 | |

| | |
|---|---|
| Total: | 41 |
| Overall Rank: | 41/19 = 2.158 |
| F1–F7: | 22/7 = 3.143 |
| F8–F13: | 12/6 = 2 |
| F14–F19: | 7/6 = 1.167 |

**Table 6: GOOSE, FDO, DA, PSO, and GA ranking on the standard benchmark functions.**

| Test Function | GOOSE | FDO | DA | PSO | GA |
|---|---|---|---|---|---|
| F1 | 4 | 1 | 2 | 3 | 5 |
| F2 | 4 | 1 | 2 | 3 | 5 |
| F3 | 3 | 1 | 2 | 4 | 5 |
| F4 | 3 | 1 | 2 | 4 | 5 |
| F5 | 3 | 2 | 1 | 4 | 5 |
| F6 | 4 | 1 | 3 | 2 | 5 |
| F7 | 1 | 5 | 3 | 2 | 4 |
| F8 | 3 | 2 | 5 | 1 | 4 |
| F9 | 1 | 3 | 4 | 2 | 5 |
| F10 | 2 | 1 | 3 | 4 | 5 |
| F11 | 1 | 4 | 3 | 2 | 5 |
| F12 | 2 | 4 | 3 | 1 | 5 |
| F13 | 3 | 4 | 1 | 2 | 5 |
| F14 | 2 | 1 | 3 | 5 | 4 |
| F15 | 1 | 2 | 5 | 4 | 3 |
| F16 | 1 | 2 | 5 | 3 | 4 |
| F17 | 1 | 2 | 5 | 3 | 4 |
| F18 | 1 | 4 | 5 | 3 | 2 |
| F19 | 1 | 2 | 4 | 5 | 3 |

**Table 7: GOOSE, FDO, DA, PSO, and GA total number of ranking on the standard benchmark functions.**

| Test Function | GOOSE | FDO | DA | PSO | GA |
|---|---|---|---|---|---|
| First | 8 | 7 | 2 | 2 | 0 |
| Second | 3 | 6 | 4 | 5 | 1 |
| Third | 5 | 1 | 6 | 5 | 2 |
| Fourth | 3 | 4 | 2 | 5 | 5 |
| Fifth | 0 | 1 | 5 | 2 | 11 |

Table 8: GOOSE, FDO, FOX, DA, BOA, WOA, and ChOA ranking on the standard benchmark functions.

| Test Function | GOOSE | FDO | FOX | DA | BOA | WOA | ChOA |
|---|---|---|---|---|---|---|---|
| F1 | 7 | 4 | 1 | 5 | 6 | 3 | 2 |
| F5 | 7 | 4 | 6 | 2 | 3 | 1 | 5 |
| F8 | 1 | 5 | 3 | 6 | 7 | 2 | 4 |
| F9 | 7 | 6 | 2 | 5 | 3 | 4 | 1 |
| F11 | 6 | 1 | 2 | 5 | 7 | 3 | 4 |

Table 9: GOOSE, SSA, DA, and WOA ranking on the modern standard benchmark functions.

| Test Function | GOOSE | SSA | DA | WOA |
|---|---|---|---|---|
| CEC01 | 4 | 1 | 3 | 2 |
| CEC02 | 4 | 2 | 3 | 1 |
| CEC03 | 1 | 3 | 4 | 2 |
| CEC04 | 4 | 1 | 2 | 3 |
| CEC05 | 4 | 1 | 2 | 3 |
| CEC06 | 1 | 2 | 3 | 4 |
| CEC07 | 1 | 2 | 4 | 3 |
| CEC08 | 1 | 2 | 3 | 4 |
| CEC09 | 2 | 1 | 4 | 3 |
| CEC10 | 1 | 2 | 3 | 4 |

Table 10: GOOSE, SSA, DA, and WOA total number of ranking on the modern benchmark functions.

| Test Function | GOOSE | SSA | DA | WOA |
|---|---|---|---|---|
| First | 5 | 4 | 0 | 1 |
| Second | 1 | 5 | 2 | 2 |
| Third | 0 | 1 | 5 | 4 |
| Fourth | 4 | 0 | 3 | 3 |

Table 11: GOOSE ranking on standard benchmark functions by type and in total.

| Test Function Type | Total Ranking | Total Ranking/No. of Function | Ranking |
|---|---|---|---|
| Unimodal | 22 | 22/7 | 2.158 |
| Multimodal | 12 | 12/6 | 2 |
| Composite | 7 | 7/6 | 1.167 |
| Total | 41 | 41/19 | 5.828 |

### 3.4 Statistical Tests

The Wilcoxon rank-sum test p values are calculated for all test functions to demonstrate that the findings reported in Tables (2), (3), and (4) are statistically significant. The results of a statistical comparison are shown in Tables (12), and (13). Because the DA algorithm has previously been reviewed against PSO and GA in this article (Mirjalili, 2016) and the FDO algorithm has already been tested against DA in this study (Abdullah & Ahmed, 2019), the research cited shows that when compared to PSO and GA, the DA outcomes are statistically significant, only the GOOSE and FDO algorithms are compared in Table 12. Moreover, the FDO algorithm has won the best result seven times, behind the GOOSE algorithm, which is eight times ahead of the ranking in this study.

The GOOSE findings are again deemed significant in all statistical tests (unimodal, multimodal, and composite test functions), as shown in Table 2, except F5, F11, F14, F15, and F18, where the values are higher than 0.05. Additionally, the comparesion test functions did not provide any unusual results. The results of the composite test functions (TF14–TF19), as given in Table 12, demonstrate that the GOOSE algorithm consistently produces results that are competitive with those of the competition. The superiority is not, however, as substantial as that of the unimodal and multimodal test functions, according to the p values. Three retain the null hypothesis in those intervals. This is because the composite test functions are tough for the methods used in this study owing to their complexity. Composite test methods measure the combined exploration and exploitation (Mirjalili, 2016). These findings demonstrate that the GOOSE algorithm's operators correctly balance exploration and exploitation to manage complexity in a tough search space. While the composite search spaces are extremely comparable to the actual search spaces, these findings make the GOOSE algorithm potentially able to address robust optimization concerns.

The p values reported in Table 13 also show that the GOOSE algorithm shows significantly better results than DA, WOA, and SSA in all statistical tests. However, the compared against the SSA algorithm provides very competitive results and outperforms, except in CEC06-CEC08, that is because the results are more than 0.05. There aren't unusual results in the modern benchmark test functions.

GOOSE outranks the other algorithms by offering a lower p-value of ranking in the majority of the situations.

Table 12: The Wilcoxon rank-sum test for classical benchmarks.

| F | GS vs. FDO |
|---|---|
| F1 | 0.000002 |
| F2 | 0.000002 |
| F3 | 0.000002 |

| F | | |
|---|---|---|
| F4 | 0.000002 | |
| F5 | **0.271155** | |
| F6 | 0.000002 | |
| F7 | 0.000002 | |
| F8 | 0.000014 | |
| F9 | 0.000002 | |
| F10 | 0.000002 | |
| F11 | **0.829013** | |
| F12 | 0.000002 | |
| F13 | 0.000053 | |
| F14 | **0.611331** | |
| F15 | **0.262173** | |
| F16 | 0.015959 | |
| F17 | 0.000002 | |
| F18 | **0.131668** | |
| F19 | 0.002981 | |

Table 13: The Wilcoxon rank-sum test for modern benchmark functions.

| F | GS vs. DA | GOOSE vs. WOA | GOOSE vs. SSA |
|---|---|---|---|
| CEC01 | 0.006424 | 0.003609 | 0.003609 |
| CEC02 | 0.000148 | 0.000002 | 0.000002 |
| CEC03 | 0.000002 | 0.00000004 | 0.00000007 |
| CEC04 | 0.000002 | 0.000003 | 0.000002 |
| CEC05 | 0.003379 | 0.000002 | 0.000002 |
| CEC06 | 0.000002 | 0.000002 | **0.503833** |
| CEC07 | 0.000002 | 0.000075 | **0.338856** |
| CEC08 | 0.000002 | 0.020671 | **0.298944** |
| CEC09 | 0.000002 | 0.000010 | 0.000002 |
| CEC10 | 0.000205 | 0.000002 | 0.000002 |

### 3.5 Quantitative Measurement Metrics

Four new measures are used in the following paragraphs to observe and analyze the performance of the suggested GOOSE algorithm in more detail. This experiment's major goals are to verify convergence and forecast how the GOOSE algorithm could behave while tackling actual issues. The positions of the goose from the first to the last iteration (search history), the value of a parameter from the first to the last iteration (trajectory), the average fitness of the goose from the first to the last iteration, and the fitness of the best score obtained from the first to the last iteration (convergence) are the quantitative metrics used. We investigate if and how the GOOSE method utilizes the search space by tracking the location of the gooses throughout optimization. During optimization, keeping an eye on a parameter's value helps us track the progression of potential solutions. The parameters should ideally alter abruptly during the exploration phase and gradually throughout the exploitation phase. The average fitness of the goose throughout optimization also demonstrates the rise in the fitness of the whole swarm. Finally, the fitness of the highest score demonstrates the growth of the realized global optimum throughout optimization.

Over a maximum of 100 iterations, 10 search agents are used to select and solve some of the functions (F2, F10, and F17). Figures 8, 9, 10, and 11 show the findings. The location of the goose over time during optimization is shown in Figure 8. One may see that the GOOSE algorithm tends to thoroughly scan the promising areas of the search space. The behavior of GOOSE while solving the composite test function F17 is intriguing since it seems that a large portion of the search area has been covered. This demonstrates the capability of GOOSE's artificial gooses to efficiently explore the search space.

Figure 9 shows the evolution of the first artificial goose's variable over 100 iterations. In the earliest iterations, it can be seen that there are sudden shifts. Throughout repetitions, these modifications eventually become less significant. This behavior may ensure that an algorithm finally converges to a point and searches locally in a search space, according to Berg et al. (Van Den Bergh & Engelbrecht, 2006).

The average fitness and convergence curves for all geese are shown in Figures 10 and 11. On every test function, the average fitness of the goose displays a diminishing behavior. This demonstrates that the GOOSE algorithm enhances the original random population's overall fitness. The convergence curves exhibit a similar pattern of behavior. This demonstrates that as the iteration counter rises, so does the accuracy of the global optimal approximation. The faster tendency in the convergence curves is another apparent feature. This is a result of the increasing focus on local search and exploitation, which greatly speeds up the convergence towards the optimum in the last stages of iterations.

In the final analysis, this section's findings demonstrated that the suggested GOOSE algorithm exhibits significant levels of exploitation and exploration.

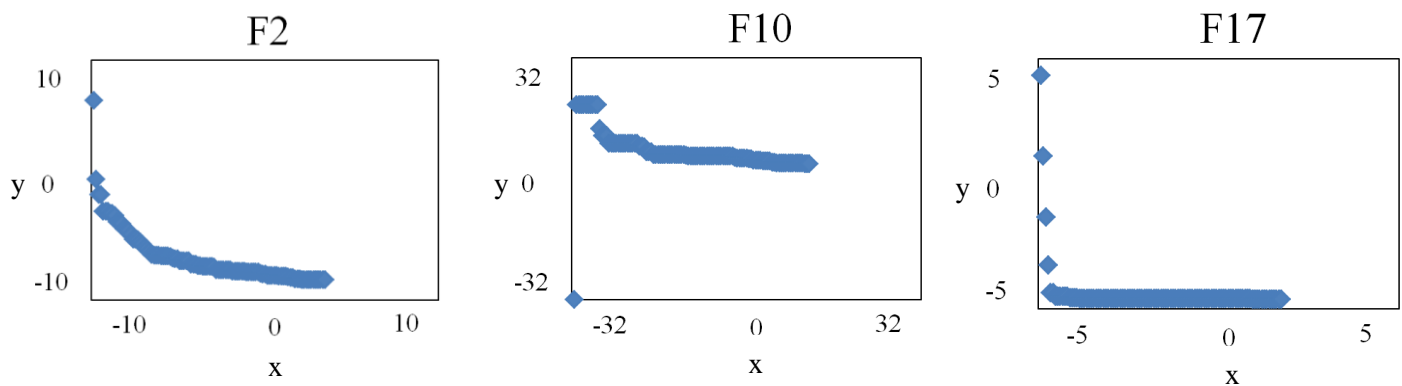

**Fig. 8: Search history of the GOOSE algorithms on unimodal, multi-modal, and composite test functions**

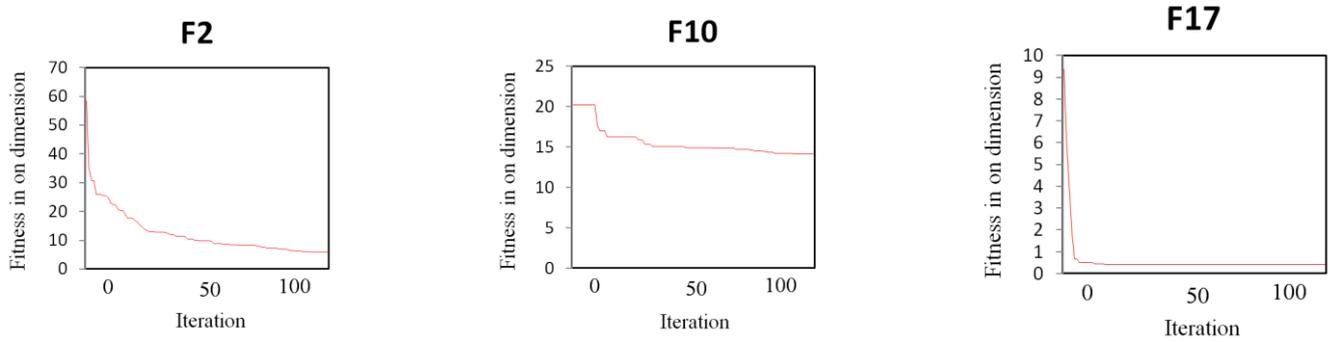

Fig. 9: Trajectory of GOOSE's search agents on unimodal, multi-modal, and composite test functions

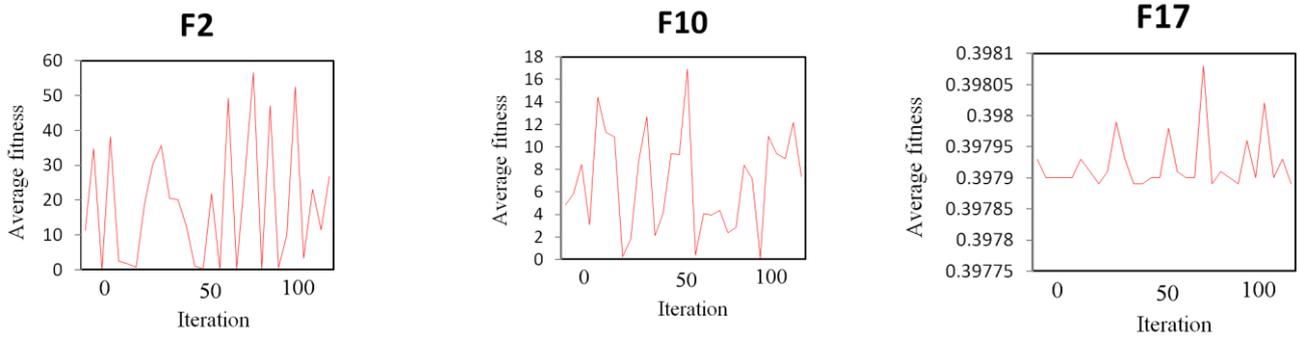

Fig. 10: Average fitness of GOOSE's search agents on unimodal, multi-modal, and composite test functions

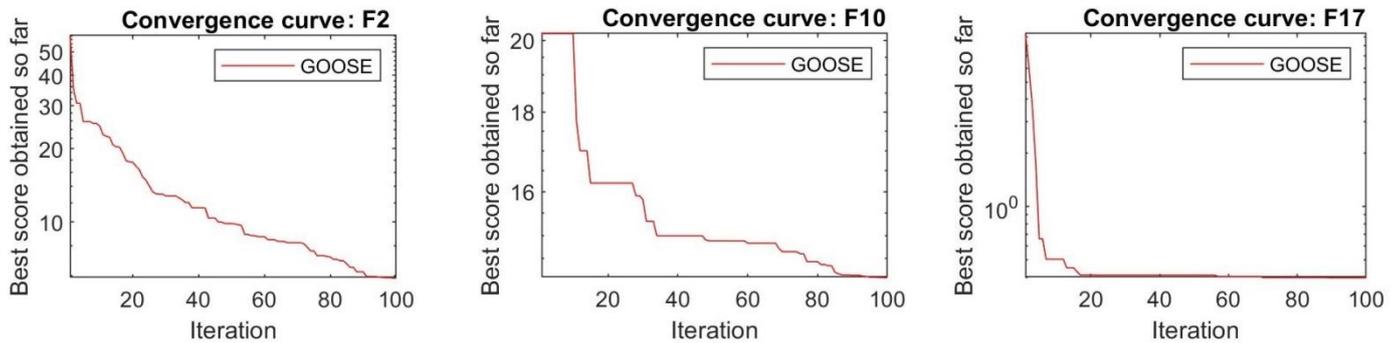

Fig. 11: Convergence curve of the GOOSE algorithms on unimodal, multi-modal, and composite test functions

### 3.6 Numerical Experiments and Implementation

The GOOSE was programmed in MATLAB R2019a and examined contrary to algorithms such as the Genetic Algorithm(GA), Particle swarm optimization(PSO), Dragonfly Algorithm(DA), and Fitness Dependent Optimizer(FDO). Also, and executed on an Intel(R) Core(TM) i7-7500U CPU @ 2.70GHz  2.90 GHz, 500 GB SSD, and 16 GB RAM.

**Table14: Parameters setting.**

| Parameters | Numbers |
| --- | --- |
| Iteration | 500 |
| Run of algorithms | 30 |
| Search agents | 30 |
| Number of dimensions | Dimension settings will be based on the use of algorithms following Tables A1–A6 in Appendix A. |

### 3.7 Setting Paremeters

The subsequent parameters were employed by the algorithms in the present paper, as shown in Table 14.

### 3.8 Algorithm Complexity

This section explains algorithm complexity, When it comes to GOOSE's computational complexity, each iteration has a timing complexity of O(SearchAgents * D * it), while SearchAgents is the population size, D is the problem dimension, and it is the number of iterations. Consequently, it may be claimed that the time complexity of GOOSE is $O(n^2)$. Furthermore, the vectors and matrices in Algorithm1 are used to determine the GOOSE space complexity. Therefore, each iteration of GOOSE has an $O(n^2)$ space complexity.

### 3.9 Real-World Applications of GOOSE

Optimization algorithms are computational procedures used to find the best solution (maximum or minimum) to a given problem. These algorithms are applied to optimization problems, which involve finding the optimal solution from a set of feasible solutions. The connection between optimization algorithms and optimization problems is fundamental and can be listed as follows:(objective function, decision variables, constraints, optimal solution, search space, optimization algorithm, iterative process, and convergence). Optimization algorithms play a key, constructive, and effective role in solving most real-life problems. In this review article, we researched several new algorithms that improve health problems (Hamad & Rashid, 2023b).

The selection of applications is based on several criteria, such as their use in different fields (engineering, science, and medicine), and these applications have long been used to test newly proposed algorithms. To demonstrate the algorithm's viability and assess its effectiveness, GOOSE has been employed to address three classical engineering and a novel application of real-world application challenges.

#### 3.9.1 Welded beam design

The first problem is taken from (Mirjalili & Lewis, 2016) and is to be developed with the least end deflection and minimum cost while adhering to the buckling load, bending stress, and shear stress limitations. Welded beam design is a critical aspect of structural engineering, which involves the creation of welded connections in beams to ensure the structural integrity and safety of various real-world structures. The welded beam construction, seen in Figure 12, is a real-world design issue that has been used as a benchmark for evaluating various optimization techniques. The goal is to determine the welded beam's least fabrication cost while taking into consideration side, and end deflection ($\delta$), buckling load ($P_c$), bending stress ($\theta$), and shear stress constraints($\tau$). h(=x1), l(=x2), t(=x3), and b(=x4) are the four design variables. The pair of objectives optimization problem is mathematically expressed as follows:

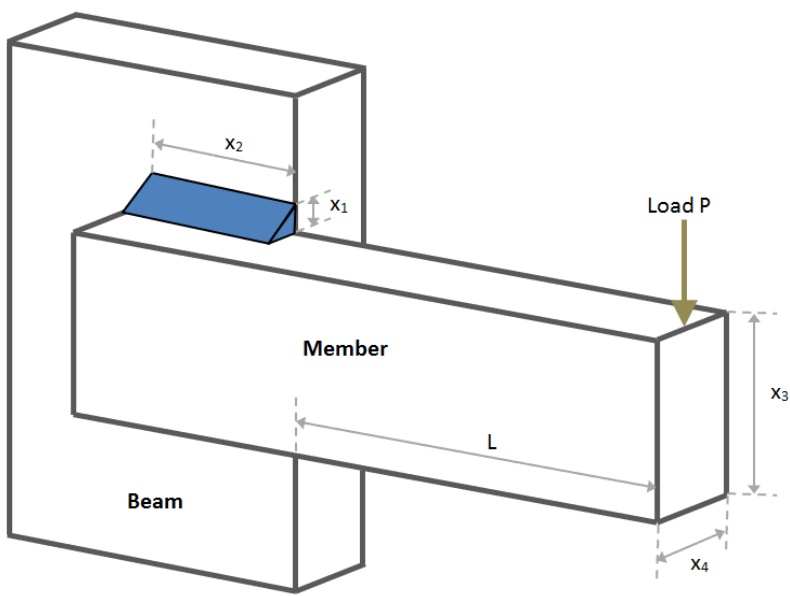

**Fig. 12: Welded beam design problem**

Consider minimising

$$\vec{x} = [x_1\ x_2\ x_3\ x_4] = [h\ l\ t\ b],$$

$$f(\vec{x}) = 1.10471 X_1^2 + 0.04811 x_3 x_4 (14.0 + x_2)$$

subject to

$$g_1(\vec{x}) = \tau(\vec{x}) - \tau_{max} \leq 0,$$

$$g_2(\vec{x}) = \sigma(\vec{x}) - \sigma_{max} \leq 0,$$

$$g_3(\vec{x}) = \delta(\vec{x}) - \delta_{max} \leq 0, \qquad (14)$$

$$g_4(\vec{x}) = x_1 - x_4 \leq 0,$$

$$g_5(\vec{x}) = P - P_c(\vec{x}) \leq 0,$$

$$g_6(\vec{x}) = 0.125 - x_1 \leq 0,$$

$$g_7(\vec{x}) = 1.10471 X_1^2 + 0.04811 x_3 x_4 (14.0 + x_2) - 5.0 \leq 0$$

Variable range

$$0.1 \leq x_1 \leq 2,$$

$$0.1 \leq x_2 \leq 10,$$

$$0.1 \leq x_3 \leq 10,$$

$$0.1 \leq x_4 \leq 2,$$

Where

$$\tau(\vec{x}) = \sqrt{(\tau')^2 + 2\tau'\tau''\frac{x_2}{2R} + (\tau'')^2},$$

$$\tau' = \frac{P}{\sqrt{2}x_1 x_2},\ \tau'' = \frac{MR}{J},\ M = P\left(L + \frac{x_2}{2}\right),$$

$$R = \sqrt{\frac{x_2^2}{4} + \left(\frac{x_1 + x_3}{2}\right)^2},$$

$$J = 2\left\{\sqrt{2}x_1 x_2 \left[\frac{x_2^2}{4} + \left(\frac{x_1 + x_3}{2}\right)^2\right]\right\},$$

$$\sigma(\vec{x}) = \frac{6PL}{x_4 x_3^2},\ \delta(\vec{x}) = \frac{6PL^3}{E x_3^2 x_4}$$

$$P_c(\vec{x}) = \frac{4.013 E \sqrt{\frac{x_3^2 x_4^6}{36}}}{L^2}\left(1 - \frac{x_3}{2L}\sqrt{\frac{E}{4G}}\right),$$

P=600lb, $L$=14 in., $\delta_{max}$=0.25 in., $E = 30 \times 1^6$ psi, G=12 x $10^6$ psi, $\tau_{max} = 13{,}600$ psi, $\sigma_{max} = 30{,}000$ psi,

Many researchers have relied on metaheuristic techniques to handle this optimization issue, among which are WOA, PSO, and GSA, as mentioned in this paper by Mirjalili and Lewis (Mirjalili & Lewis, 2016). According to the optimization findings shown in Table 15, GOOSE converged to the third-best design. The statistical findings from 30 different runs of various algorithms. To solve this issue, we used 20 search agents and a maximum number of 500 iterations of the search. This table demonstrates that GOOSE once more performs better on average.

**Table 15: Comparison of GOOSE statistical results with literature for the welded beam design problem** (Mirjalili & Lewis, 2016).

| Algorithm | Average | Standard deviation |
| --- | --- | --- |
| WOA | 1.7320 | 0.0226 |
| PSO | 1.7422 | 0.01275 |
| GOOSE | 3.1882 | 0.03996 |
| GSA | 3.5761 | 1.2874 |

### 3.9.2 Economic Load Dispatch Problem

The second problem is the economic load dispatch problem. The Economic Load Dispatch (ELD) problem is a significant issue in power system operations and has real-world implications. Its primary objective is to allocate the optimal power output to various generating units in a power grid to meet the electricity demand while minimizing the total operating cost. Accordingly, minimum fuel should be used to optimize the power generation unit to lower the operation cost of generating energy. Equation (15) illustrates the function of ELD. The economic load dispatch problem is an optimization problem in the field of electricity. The primary objective is to minimize the energy production cost while taking into account the load demand within various equality and inequality constraints (Pradhan et al., 2018), (Nischal & Mehta, 2015).

$$C_t = \sum_{i=1}^{n} C_i(P_i) \qquad (15)$$

Where $C_i$ is the needed cost by generator, $P_i$ is the real power generated by generator $i$, $C_t$ is the total cost of fuel, and $n$ is the number of generators. Therefore, the following quadratic function and Equation (16) must be improved to represent $C_i$:

$$C_i = \sum_{i=1}^{n} a_i P_i^2 + b_i P_i + C_i \qquad (16)$$

where $a_i$, $b_i$, and $C_i$ are each utilized as a generator $i$'s coefficient costs. Two conditions must be met for the aforementioned equation to be valid: the power generator must not exceed its capacity and the aggregate of all power generators must meet the power demand with power loss (Sharma et al., 2017), (Kamboj et al., 2016).

To satisfy the 150 MW power needs and tackle the problem of economical load dispatch using three power generators, 50 separate runs and 1000 iterations of GOOSE were used to find the solution to this problem. Table 16 lists the data from the three distinct generators. Subsequently, the GOOSE results were compared with those of GWO, PSO, WOA, FDO, and FOX. All these techniques were tested against the GOOSE algorithm, as referenced in (Mohammed and Rashid, 2022). The findings demonstrated that, in terms of generating power, FOX, PSO, GOOSE and WOA had the same result, GWO and FDO, in this order, produce the best power output. At the same time, in terms of product cost, the goose algorithm costs more than the other algorithms. Table 17 demonstrates the GOOSE results compared to the other methods.

Table 16: Economic load dispatch for 3-generating-unit system (Load demand = 150 MW)

| Unit / Algorithm | Power | Cost |
|---|---|---|
| P1 (MW) | 45 | 608.06 |
| P2 (MW) | 57.5 | 1118 |
| P3 (MW) | 47.5 | 761.89 |
| Total Generated Power(MW) | 150 | 2487.95 |

Table 17: Comparison of GOOSE with other Algorithms for Three Generators ELD Problem (Mohammed & Rashid, 2022).

| Unit / Algorithm | GWO | PSO | WOA | FDO | FOX | GOOSE |
|---|---|---|---|---|---|---|
| P1 (MW) | 31.94 | 60.0345 | 31.938 | 32.665 | 31.937 | 45 |
| P2 (MW) | 67.284 | 25.6626 | 67.284 | 65.489 | 67.277 | 57.5 |
| P3 (MW) | 50.777 | 67.2313 | 50.778 | 51.846 | 50.785 | 47.5 |
| Total Generated Power(MW) | 150.001 | 152.9285 | 150 | 149.999 | 152.6089 | **150** |
| Cost($/hr) | 1579.698 | 1637.084 | 1579.699 | 1579.87 | 1579.699 | 2487.95 |

### 3.9.3 The Pathological IgG Fraction in the Nervous System

The third problem that was selected was the Pathological IgG. Aladdin and Rashid proposed a new application, known as "The Pathological IgG Fraction in the Nervous System"(Aladdin & Rashid, 2023) The goal of this topic is to find the best clarify for the features of the most effective evaluation of pathological IgG levels in CSF induced to highlight the nervous system fluctuation. According to Equation (18), which is improved from the collection of statistical regression lines (LEFVERT & LINK, 1985), (Su & Chiu, 1986), the frequency of the regression line passing through the origin is appropriate for statistical and functional reasons. The majority of the research aimed to determine a relationship between serum and fluid albumin concentrations.

After analyzing the IgG quotient for the patient's specific albumin ratio, Equation (17) (Su & Chiu, 1986)may be utilized to calculate the locally produced concentration of pathological *(IgGp)* in CSF. The confidence interval of the IgG quotient (y) for a certain albumin quotient (*x*) is provided by these two variables, and *STD(x,y)* is the standard deviation of the (*y*) values from the regression line between (-0.001, +0.001).

$$IgGp = IgG(CSF) - (0.43\, Alb(Serum) - Alb(CSF) + 0.001) * IgG(Serum) \tag{17}$$

To establish that

$$IgG = X_i \quad \text{So, } IgG(IgGp) = Y(X_i)$$

$$Y(X_i) = \sum_{i=1}^{n}(0.41 + 0.0014X_i) \tag{18}$$

The GOOSE method is used to optimize this issue while keeping in mind the restrictions of Equation (18).150 iterations are performed using 12 search agents. The result is shown in Figure 13 and includes both the average fitness value and the global average fitness for each iteration. The study shows that the globally optimized solution's iteration 135 gave the best outcome, which is (0.00047792). In the same situation, the LEO Algorithm obtained (5.088). Thus, the result obtained shows that the GOOSE Algorithm performs effectively.

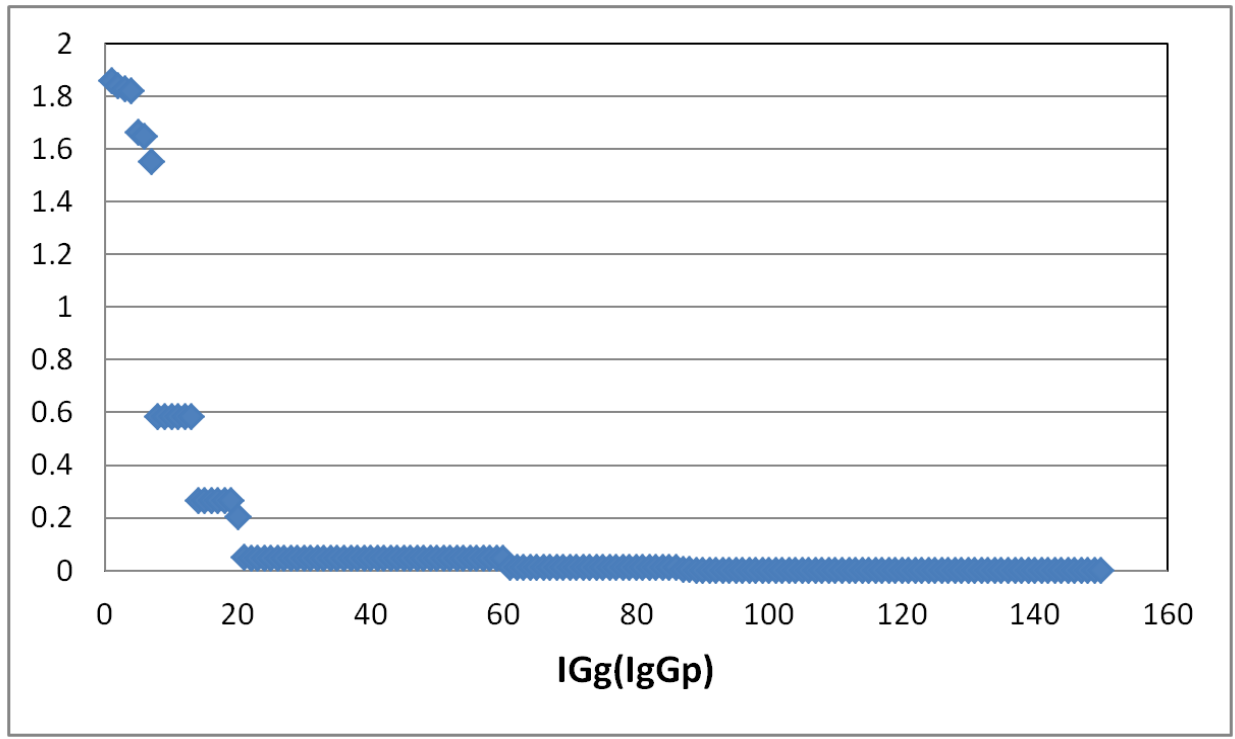

**Fig. 13: Best overall fitness scores from 150 iterations with 12 search agents in the nervous system's (IgGp)**

### 3.9.4 Pressure Vessel Design Problem

The final problem that was selected was the Pressure Vessel Design Problem. The author of the original proposal, Sandgren (Sandgren, 1990), provided a description of the pressure vessel design problems. Pressure vessels, which provide dependable storage for pressurized liquids and gases, are essential components of many production facilities and processing plants. As shown in Figure. 14, a pressure vessel is often built as a cylinder surrounded by hemispherical end caps or heads. The objective of this task was to reduce the overall cost of producing this structure by considering the expenses of materials, shaping, and welding. Four continuous variables and four inequality constraints are included in the design problem.

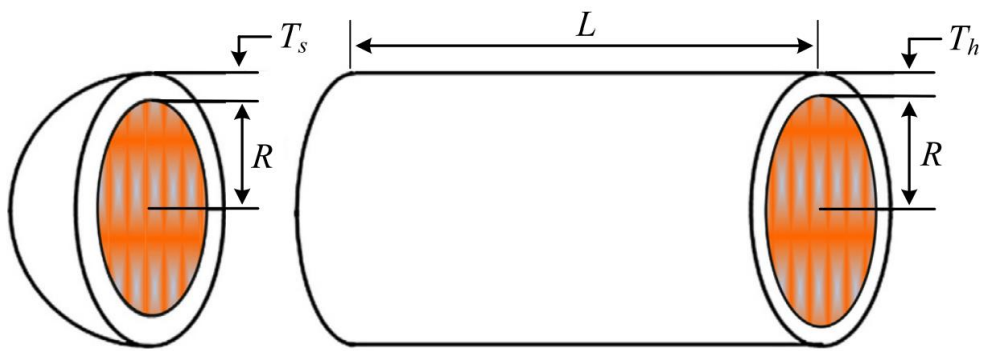

**Fig. 14: Visualization of pressure vessel design problem.**

Four design factors need to be optimized in this problem:
- Thickness of the shell ($T_s$)
- Thickness of the head ($T_h$)
- Inner radius ($R$)
- Length of the cylindrical section of the vessel ($L$)

Of these variables, the lower and upper limits are shown below for the discrete values $T_s$ and $T_h$, which are of the order of integer multiplies of 0.0625, and the continuous values $R$ and $L$.

$$1 \times 0.0625 \leq T_s, \quad T_h \leq 99 \times 0.0625, \quad 10.0 \leq R \leq 200.0, \quad 10.0 \leq L \leq 200.0 \tag{19}$$

The four inequality restrictions and the problem formulation can be expressed mathematically as follows:

Minimize
$$f(T_s, T_h, R, L) = 0.6224 T_s R L + 1.7781 T_h R^2 + 3.1661 T_s^2 L + 19.84 T_s^2 R \qquad (20)$$

Subject to
$$\begin{cases} g_1 = -T_s + 0.0193 R \leq 0 \\ g_2 = -T_h + 0.0095 R \leq 0 \\ g_3 = -\pi R^2 L - \frac{4}{3}\pi R^3 + 1296000 \leq 0 \\ g_4 = L - 240 \leq 0 \end{cases} \qquad (21)$$

Researchers have addressed the pressure vessel design issue in several studies, using it as a common benchmark. The results are shown in Table 18 and compared with other algorithms such as CEPSO, PSO, MBA, GWO, QOCSOS, SFS, and ISOS referenced from (Çelik, 2020a) and mSSA referenced from (Çelik et al., 2021).

**Table 18: Comparison of GOOSE statistical results with literature for the pressure vessel design problem** (Çelik, 2020a) **and** (Çelik et al., 2021)**.**

| Algorithm | Best |
|---|---|
| CEPSO | 6061.0777 |
| PSO | 6059.7143 |
| MBA | 5889.3216 |
| GWO | 6051.5639 |
| QOCSOS | 5885.332774 |
| SFS | 6059.714335 |
| ISOS | 6059.714335 |
| mSSA | 6059.71433 |
| GOOSE | 6343.6587 |

## 4. Conclusions

In summary, our work introduced a novel meta-huristic optimization technique that was motivated by Goose guarding behavior. The suggested approach is also known as the GOOSE Optimization Algorithm. To examine the exploration, exploitation, local optima avoidance, and convergence behavior of the proposed algorithm, a thorough analysis of 19 common benchmark functions, 10 benchmark functions from current mathematics, and 5 benchmark functions from classical mathematics was undertaken. GOOSE was discovered to be sufficiently competitive with other cutting-edge meta-heuristic techniques. Additionally, all stages of searching, including initialization, exploration, and exploitation, rely on the randomization process of GOOSE.

The findings demonstrated that GOOSE was able to provide results that were very competitive with those of well-known heuristics including FDO, DA, PSO, GA, FOX, BOA, WOA, ChOA, and SA. First, the findings for the unimodal functions demonstrated the GOOSE algorithm's underutilization. Second, the findings on multimodal functions supported GOOSE's capacity for exploration. Third, the composite functions' findings demonstrated greater high local optima avoidance.

Moreover, the outcomes of the engineering design issues demonstrated that the GOOSE algorithm performs well in uncharted, difficult search regions. The GOOSE method was eventually used to solve a genuine optical engineering issue. The findings on this topic demonstrated a significant improvement over existing methods, demonstrating the relevance of the suggested strategy in resolving actual issues. For future works, several research directions can be recommended. Hybridizing other algorithms with GOOSE and modifying the GOOSE.

**Appendix A:** Benchmark test functions

**Table A1: Unimodal benchmark functions** (Marcin Molga, 2005).

| Functions | Dim | Range | $f$min |
|---|---|---|---|
| $f_1(x) = \sum_{i=1}^{n} X_i^2$ | 10 | [-100,100] | 0 |
| $f_2(x) = \sum_{i=1}^{n} \|x\| + \prod_{i=1}^{n} = 1\|x_i\|$ | 10 | [-10,10] | 0 |
| $f_3(x) = \sum_{i=1}^{n} \left(\sum_{j-1}^{i} x_j\right)^2$ | 10 | [-100,100] | 0 |
| $f_4(x) = max_i\{\|x_i\|, 1 \leq i \leq n\}$ | 10 | [-100,100] | 0 |
| $f_5(x) = \sum_{i=1}^{n-1}[100(x_{i+1} - x_i^2)^2 + (x_i - 1)^2]$ | 10 | [-30,30] | 0 |
| $f_6(x) = \sum_{i=1}^{n}([x_i + 0.5])^2$ | 10 | [-100,100] | 0 |
| $f_7(x) = \sum_{i=1}^{n} ix_i^4 + random[0,1]$ | 10 | [-1.28,1.28] | 0 |

**Table A2: Multimodal benchmark functions** (Marcin Molga, 2005).

| Functions | Dim | Range | $f$min |
|---|---|---|---|
| $f_8(x) = \sum_{i=1}^{n} -x_i \sin(\sqrt{\|x_i\|})$ | 10 | [-500,500] | -418.9829 |
| $f_9(x) = \sum_{i=1}^{n} \|x\| + \prod_{i=1}^{n} = 1\|x_i\|$ | 10 | [-5.12,5.12] | 0 |
| $f_{10}(x) = -20exp\left(-0.2\sqrt{\frac{1}{n}\sum_{i=1}^{n} x_i^2}\right) - exp\left(\frac{1}{n}\sum_{i=1}^{n} \cos(2\pi x_i)\right) + 20 + e$ | 10 | [-32,32] | 0 |
| $f_{11}(x) = \frac{1}{4000}\sum_{i=1}^{n} x_i^2 - \prod_{i=1}^{n} \cos\left(\frac{x_i}{\sqrt{i}}\right) + 1$ | 10 | [-600,600] | 0 |
| $f_{12}(x) = \frac{\pi}{n}\{10 \sin(\pi y_1) + \sum_{i=1}^{n-1}(y_i - 1)^2[1 + 10sin^2(\pi x_i)] + (y_n - 1)^2\} + \sum_{i=1}^{n} u(x_i, 10, 100, 4)$<br>$y_i = 1 + \frac{x_i+1}{4}$<br>$u(x_i, a, k, m) = \begin{cases} k(x_i - a)^m & x_i > a \\ 0 & -a < x_i < a \\ k(-x_i - a)^m & x_i < -a \end{cases}$ | 10 | [-50,50] | 0 |
| $f_{13}(x) =$<br>$0.1\{sin^2(3\pi x_1) + \sum_{i=1}^{n}([x_i - 1])^2[1 + sin^2(3\pi x_i + 1)] + (x_n - 1)^2 + [1 + sin^2(2\pi x_n)]\} +$<br>$\sum_{i=1}^{n} u(x_i, 10, 100, 4)$ | 10 | [-50,50] | 0 |
| $f_{14}(x) = -\sum_{i=1}^{n} \sin(x_i) \cdot \left(\sin\left(\frac{ix_i^2}{\pi}\right)\right)^{2m}, m = 10$ | 10 | [0,π] | -4.687 |
| $f_{15}(x) = \left[e^{-\sum_{i=1}^{n}(x_i/\beta)^{2m}} - 2e^{\sum_{i}^{n} x_i^2}\right] \cdot \prod_{i=1}^{n} cos^2 x_i, m = 5$ | 10 | [-20,20] | -1 |

| Functions | | | Dim | Range | $f$min |
|---|---|---|---|---|---|
| $f_{16}(x) = \{[\sum_{i=1}^{n} sin^2(x_i)] - exp(-\sum_{i=1}^{n} x_i^2)\} \cdot exp[-\sum_{i=1}^{n} sin^2 \sqrt{|x_i|}]$ | | | 10 | [-10,10] | -1 |

**Table A3: Fixed-dimension multimodal benchmark functions** (Marcin Molga, 2005).

| Functions | Dim | Range | $f$min |
|---|---|---|---|
| $f_{14}(x) = \left(\frac{1}{500} + \sum_{j=1}^{25} \frac{1}{j+\sum_{i=1}^{2}(x_i - a_{ij})^6}\right)^{-1}$ | 2 | [-65,65] | 1 |
| $f_{15}(x) = \sum_{i=1}^{11} \left[a_i - \frac{x_1(b_i^2 + b_i x_i)}{b_i^2 + b_i x_3 + x_4}\right]^2$ | 4 | [-5,5] | 0.0003 |
| $f_{16}(x) = 4x_1^2 - 2.1x_1^4 + \frac{1}{3}x_1^6 + x_1 x_2 - 4x_2^2 + 4x_2^4$ | 2 | [-5,5] | -1.0316 |
| $f_{17}(x) = \left(x_2 - \frac{5.1}{4\pi^2}x_1^2 + \frac{5}{\pi}x_1 - 6\right)^2 + 10\left(1 - \frac{1}{8\pi}\right)\cos x_1 + 10$ | 2 | [-5,5] | 0.398 |
| $f_{18}(x) = [1 + (x_1 + x_2 + 1)^2(14 - 19x_1 + 3x_1^2 - 14x_2 + 6x_1 x_2 + 3x_2^2)]x[30 + (2x_1 - 3x_2)^2 x(18 - 32x_2 + 12x_1^2 + 48x_2 - 36x_1 x_2 + 27x_2^2)]$ | 2 | [-2,2] | 3 |
| $f_{19}(x) = -\sum_{i=1}^{4} c_i \exp\left(-\sum_{j=1}^{3} a_{ij}(x_j - p_{ij})^2\right)$ | 3 | [0,1] | -3.86 |
| $f_{20}(x) = -\sum_{i=1}^{4} c_i \exp\left(-\sum_{j=1}^{6} a_{ij}(x_j - p_{ij})^2\right)$ | 6 | [0,1] | -3.32 |
| $f_{21}(x) = -\sum_{i=1}^{5} [(X - a_i)(X - a_i)^T + c_i]^{-1}$ | 4 | [0,10] | -10.1532 |
| $f_{22}(x) = -\sum_{i=1}^{7} [(X - a_i)(X - a_i)^T + c_i]^{-1}$ | 4 | [0,10] | -10.4028 |
| $f_{23}(x) = -\sum_{i=1}^{10} [(X - a_i)(X - a_i)^T + c_i]^{-1}$ | 4 | [0,10] | -10.5363 |

**Table A4: Composited Benchmarks Test Functions** (Marcin Molga, 2005).

| Functions | Dim | Range | $f$min |
|---|---|---|---|
| $F_{24}(CF1)$: <br> $f_1, f_2, f_3, \ldots, f_{10}$ =Sphere Function <br> $[\partial_1, \partial_2, \partial_3, \ldots, \partial_{10}] = [1,1,1, \ldots ,1]$ <br> $[\lambda_1, \lambda_1, \lambda_1, \ldots, \lambda_1,] = [5/100, 5/100, 5/100, \ldots, 5/100]$ | 10 | [-5,5] | 0 |
| $F_{25}(CF2)$: <br> $f_1, f_2, f_3, \ldots, f_{10}$ =Griewank's Function <br> $[\partial_1, \partial_2, \partial_3, \ldots, \partial_{10}] = [1,1,1, \ldots ,1]$ <br> $[\lambda_1, \lambda_2, \lambda_3, \ldots, \lambda_{10},] = [5/100, 5/100, 5/100, \ldots, 5/100]$ | 10 | [-5,5] | 0 |
| $F_{26}(CF3)$: <br> $f_1, f_2, f_3, \ldots, f_{10}$ =Griewank's Function <br> $[\partial_1, \partial_2, \partial_3, \ldots, \partial_{10}] = [1,1,1, \ldots ,1]$ <br> $[\lambda_1, \lambda_2, \lambda_3, \ldots, \lambda_{10},] = [1,1,1, \ldots ,1]$ | 10 | [-5,5] | 0 |
| $F_{27}(CF4)$: <br> $f_1, f_2$ =Ackley's Function <br> $f_3, f_4$ =Rastrigin's Function <br> $f_5, f_6$ =Weierstras's Function <br> $f_7, f_8$ =Griewank's Function <br> $f_9, f_{10}$ =Sphere's Function <br> $[\partial_1, \partial_2, \partial_3, \ldots, \partial_{10}] = [1,1,1, \ldots ,1]$ <br> $[\lambda_1, \lambda_2, \lambda_3, \ldots, \lambda_{10},] = [5/32, 5/32, 1,1, 5/0.5, 5/0.5, 5/100, 5/100, 5/100, 5/100]$ | 10 | [-5,5] | 0 |
| $F_{28}(CF5)$: | 10 | [-5,5] | 0 |
| $f_1, f_2$ = Rastrigin's Function<br>$f_3, f_4$ = Weierstras's Function<br>$f_5, f_6$ = Griewank's Function<br>$f_7, f_8$ = Ackley's Function<br>$f_9, f_{10}$ = Sphere's Function<br>$[\partial_1, \partial_2, \partial_3, \ldots, \partial_{10}] = [1,1,1, \ldots, 1]$<br>$[\lambda_1, \lambda_2, \lambda_3, \ldots, \lambda_{10},] = [1/5, 1/5, 5/0.5, 5/0.5, 5/100, 5/100, 5/32, 5/32, 5/100, 5/100]$ | | | |
| $F_{29}(CF6)$:<br>$f_1, f_2$ = Rastrigin's Function<br>$f_3, f_4$ = Weierstras's Function<br>$f_5, f_6$ = Griewank's Function<br>$f_7, f_8$ = Ackley's Function<br>$f_9, f_{10}$ = Sphere's Function<br>$[\partial_1, \partial_2, \partial_3, \ldots, \partial_{10}] = [0.1, 0.2, 0.3, 0.4, 0.5, 0.6, 0.7, 0.8, 0.9, 1]$<br>$[\lambda_1, \lambda_2, \lambda_3, \ldots, \lambda_{10},] =$<br>$[0.1*1/5, 0.2*1/5, 0.3*5/0.5, 0.4*$<br>$5/0.5, 0.5*5/100, 0.6*5/100, 0.7*5/32, 0.8*5/32, 0.9*5/100, 1*5/1]$ | 10 | [-5,5] | 0 |

**Table A5: CEC-C06 2019 Benchmarks "The 100-Digit Challenge:"** (Brest et al., 2019).

| Functions | Dim | Range | fmin |
|---|---|---|---|
| STORN'S CHEBYSHEV POLYNOMIAL FITTING PROBLEM | 9 | [-8192,8192] | 1 |
| INVERSE HILBERT MATRIX PROBLEM | 16 | [-16384,16384] | 1 |
| LENNARD-JONES MINMUM ENERGY CLUSTER | 18 | [-4,4] | 1 |
| RASTRIGIN'S FUNCTION | 10 | [-100,100] | 1 |
| GRIEWANK'S FUNCTION | 10 | [-100,100] | 1 |
| WEIERSTRASS FUNCTION | 10 | [-100,100] | 1 |
| MODIFIED SCHWEFEL'S FUNCTION | 10 | [-100,100] | 1 |
| EXPANDED SCHAFFER'S F6 FUNCTION | 10 | [-100,100] | 1 |
| HAPPY CAT FUNCTION | 10 | [-100,100] | 1 |
| ACKLEY FUNCTION | 10 | [-100,100] | 1 |

**Table A6: Five Classical benchmark functions.**

| Functions | Dim | Range | $f$min |
|---|---|---|---|
| Unimodal Benchmark Functions | | | |
| $f_1(x) = \sum_{i=1}^{n} x_i^2$ | 30 | [-100,100] | 0 |
| $f_5(x) = \sum_{i=1}^{n-1}[100(x_{i\_1} - x_i^2)^2 + (x_i - 1)^2]$ | 30 | [-30,30] | 0 |
| Multimodal Benchmark Functions | | | |
| $f_8(x) = \sum_{i=1}^{n} -x_i \sin(\sqrt{|x_i|})$ | 30 | [-500,500] | -418.9829 |
| $f_9(x) = \sum_{i=1}^{n}[x_i^2 - 10\cos(2\pi x_i) + 10]$ | 30 | [-5.12,5.12] | 0 |
| $f_{11}(x) = \frac{1}{4000}\sum_{i=1}^{n} x_i^2 - \prod_{i=1}^{n} \cos\left(\frac{x_i}{\sqrt{i}}\right) + 1$ | 30 | [-600,600] | 0 |

**Conflict of interest:** The authors declare no conflict of interest to any party.

**Ethical Approval:** The manuscript is conducted in the ethical manner advised by the targeted journal.

**Consent to Participate:** Not applicable

**Consent to Publish:** The research is scientifically consented to be published.

**Funding:** The research did not receive specific funding but we want to contribute to a traditional publishing model.

**Competing Interests:** The authors declare no conflict of interest.

**Availability of data and materials:** Data can be shared upon request from the corresponding author.

**Acknowledgment:** A preprint copy is available on Cornell University Server see the link below:

https://arxiv.org/abs/2307.10420.